\documentclass[lettersize,journal]{IEEEtran}
\usepackage{amsmath,amsfonts}
\usepackage{algorithmic}
\usepackage{algorithm}
\usepackage{array}
\usepackage[caption=false,font=normalsize,labelfont=sf,textfont=sf]{subfig}
\usepackage{textcomp}
\usepackage{stfloats}
\usepackage{url}
\usepackage{verbatim}
\usepackage{graphicx}
\usepackage{cite}
\usepackage[T1]{fontenc} 
\usepackage{amsmath}
\usepackage{amsthm}
\usepackage{amssymb}
\usepackage{setspace}
\usepackage{enumitem}
\usepackage{booktabs} 
\usepackage{tabularx} 
\usepackage{multirow} 
\usepackage{threeparttable}

\begin{document}

\title{Running-time Analysis of ($\mu+\lambda$) Evolutionary Combinatorial Optimization Based on Multiple-gain Estimation}
\author{Min Huang, Pengxiang Chen, Han Huang*, Tongli He, Yushan Zhang, Zhifeng Hao}
\maketitle
\begin{abstract}
	The running-time analysis of evolutionary combinatorial optimization is a fundamental topic in evolutionary computation. However, theoretical results regarding the $(\mu+\lambda)$ evolutionary algorithm (EA) for combinatorial optimization problems remain relatively scarce compared to those for simple pseudo-Boolean problems. This paper proposes a multiple-gain model to analyze the running time of EAs for combinatorial optimization problems. The proposed model is an improved version of the average gain model, which is a fitness-difference drift approach under the sigma-algebra condition to estimate the running time of evolutionary numerical optimization. The improvement yields a framework for estimating  the expected first hitting time of a stochastic process in both average-case and worst-case scenarios.
    It also introduces novel running-time results of evolutionary combinatorial optimization, including two tighter time complexity upper bounds than the known results in the case of ($\mu+\lambda$) EA for the knapsack problem with favorably correlated weights, a closed-form expression of time complexity upper bound in the case of ($\mu+\lambda$) EA for general $k$-MAX-SAT problems and a tighter time complexity upper bounds than the known results in the case of ($\mu+\lambda$) EA for the traveling salesperson problem. Experimental results indicate that the practical running time aligns with the theoretical results, verifying that the multiple-gain model is an effective tool for running-time analysis of ($\mu+\lambda$) EA for combinatorial optimization problems.
\end{abstract}


\section{Introduction}
\IEEEPARstart{I}{n} the past two decades, great advancements have been made in running-time analysis within the field of evolutionary computation. These running-time analyses provide rigorous mathematical insights into the search behaviors of evolutionary algorithms (EAs), help explain why EAs can solve problems quickly, and improve the efficiency of algorithms\cite{doerr2021survey}.

The objective of running-time analysis for EAs is to determine the number of fitness evaluations required for an algorithm to identify at least one global optimum for the first time. The first hitting time (FHT) is a commonly applied concept for estimating the running time of EAs, reflecting the number of iterations required for an EA first to attend the global optimum \cite{he2016average}. In addition, the expected first hitting time (EFHT) refers to the average number of iterations that an EA needs to attend the global optimum initially \cite{chen2009new}. Therefore, the aim of running-time analysis can be described as analyzing the EFHT of EAs.

The research on the running-time analysis of EAs is primarily divided into evolutionary numerical optimization and evolutionary combinatorial optimization. As a wide range of engineering tasks are combinatorial optimization problems, studying evolutionary combinatorial optimization is crucial. Consequently, this paper considers the EFHT of evolutionary combinatorial optimization. However, due to the randomness of EAs and the fluctuating fitness values of combinatorial optimization problems, conducting running-time analysis on evolutionary combinatorial optimization is challenging.

The running-time analysis of evolutionary combinatorial optimization has expanded from simple pseudo-Boolean functions to combinatorial optimization problems even NP. For example, Frank et al. \cite{neumann2018running-time, neumann2019running-time, xie2021running-time} analyzed the running time of solving specific instances of the knapsack problem using (1+1) EA, GSEMO, and RLS. Zhou et al.\cite{lai2013(1+1),zhou2014(1+1),zhou2015(1+1)} conducted approximation performance analyses for specific instances of the minimum label spanning tree problem, multiprocessor scheduling problem, and maximum cut problem. Doerr \cite{doerr2015improved} proved the expected running time in the case of (1+1) EA for high-density satisfiable 3-CNF formulas. Shi et al. \cite{shi2021runtime} analyzed the expected running time in the case of (1+1) EA and RLS for dynamic weighted vertex cover problem. Since the above studies have primarily focused on simplified problem instances, there remains a lack of running-time analysis results concerning more complicated EAs for combinatorial optimization problems.

Additionally, the research on the running-time analysis of evolutionary combinatorial optimization also has extended from single-objective optimization to multi-objective optimization. For example, Benjamin et al. \cite{doerr2023first, zheng2023running-time} analyzed the time complexity of NSGA-II on the OneMinMax and OneJumpZeroJump functions. Lu et al. \cite{lu2024towards} further demonstrated that the running times of R-NSGA-II on the OneJumpZeroJump and OneMinMax functions are asymptotically more efficient than the NSGA-II. Bian et al. \cite{bian2024archive,bian2023stochastic} reduced the running times of NSGA-II and SMS-EMOA on certain problems.   Qian et al. \cite{qian2024quality} demonstrated the time complexity of quality-diversity algorithm MAP-Elites on monotone approximately submodular maximization with a size constraint, and on the set cover problem. Dang et al. \cite{dang2023analysing} analyzed the time complexity of NSGA-II on LeadingOnesTrailingZeroes under a specific noise model. Cerf et al. \cite{cerf2023first} conducted the running-time analysis of NSGA-II for the bi-objective minimum spanning tree problem. Opris et al. \cite{opris2024runtime} presented the running-time analyses of NSGA-III for $m$-LOTZ, $m$-OMM, and $m$-COCZ, where $m$ represents the number of objectives. Although the above studies have yielded substantial theoretical advancements, most investigations have concentrated on basic benchmark functions.

In summary, the running-time analysis of EAs in single-objective optimization mainly focuses on simplified algorithms such as (1+1) EA and RLS. Research findings on more complex algorithms such as the $(\mu+\lambda)$ EA remain relatively scarce. This phenomenon can be attributed to the analytical challenges caused by the population-based characteristics of complicated algorithms and the fluctuating fitness values in most combinatorial problems.

With the advancement of theoretical research on EAs, a multitude of methods have been proposed for analyzing the EFHT of EAs, such as drift analysis \cite{2001Drift,chen2009new,oliveto2011simplified,MarkovChain,he2016average,fajardo2023runtime}, switch analysis \cite{yu2014switch,yu2015switch}, average gain model \cite{hhan2014,yushan2016first,huang2019experimental}, etc.
One of the most difficult problems lies in estimating the expected progress of fitness values during iterations. To address the difficulty of estimating the expected progress of fitness, Huang et al. \cite{huang2019experimental} introduced an experimental approach for estimating the EFHT of evolutionary numerical optimization based on the average gain model. The average gain was introduced in \cite{hhan2014} which is a fitness-difference pointwise drift under the sigma-algebra condition. The experimental method in \cite{huang2019experimental} utilizes surface fitting techniques to simplify analysis by replacing complex mathematical calculations. Since the surface fitting function designed in this work relies on continuity, this experimental method cannot be directly applied to evolutionary combinatorial optimization. 



Because the average gain model in \cite{yushan2016first} is insufficient to characterize the fluctuating fitness values in combinatorial optimization problems, we propose a multiple-gain model to estimate the average-case upper bounds of EFHT for evolutionary combinatorial optimization. Furthermore, since the method in \cite{huang2019experimental} is limited to evolutionary numerical optimization, we integrate the multiple-gain model with the law of large numbers to estimate the worst-case upper bounds of EFHT for evolutionary combinatorial optimization.
Because of the scarcity of theoretical results of population-based EAs for combinatorial optimization problems, we analyze the EFHT in the case of $(\mu+\lambda)$ EA for three combinatorial optimization problems. Our contributions are listed below.
\begin{itemize} 
    \item Improving the average gain model to extend its applicability to the running-time analysis of evolutionary combinatorial optimization.
	\item Providing two tighter average-case upper bounds of EFHT under certain conditions than the known results \cite{neumann2018running-time} in the case of ($\mu+\lambda$) EA for the knapsack problem with favorably correlated weights. Indeed, the conclusion in \cite{neumann2018running-time} represents a special case of our result when $\mu=1$, $\lambda=1$.
	\item Providing a closed-form expression for the average-case upper bound of EFHT in the case of a class of ($\mu+\lambda$) EA for general $k$-MAX-SAT problems. Previously, specific instances of the $k$-MAX-SAT problem have been investigated in \cite{buzdalov2017running-time,doerr2015improved,sutton2012parameterized}. This paper attempts to analyze the running time of EAs for the general $k$-MAX-SAT problem. 
        \item Providing a tighter average-case upper bound of EFHT than the known results \cite{sutton2014tsp} in the case of ($\mu+\lambda$) EA for TSP problem where every city is a vertex on its convex hull. 
    
\end{itemize}

The remainder of this article is structured as follows. Section II presents a review of running-time analysis results of various EAs. Section III presents a detailed description of the multiple-gain model. In Section IV, we utilize the multiple-gain model to estimate the average-case upper bounds of EFHT for three evolutionary combinatorial optimization instances. Section V presents experimental results of the average-case and worst-case upper bounds of EFHT for these three instances. Section VI summarizes the content of this article.

\section{Related Works}
Current studies on the running-time analysis of evolutionary combinatorial optimization primarily focus on $(1+1)$ EA, $(1+\lambda)$ EA, $(\mu+1)$ EA, and $(\mu+\lambda)$ EA.
\subsection{$(1+1)$ EA}
In the (1+1) EA, both the parent and offspring populations consist of one individual.

For the simple pseudo-Boolean problem, Droste et al. \cite{droste1998(1+1),droste2002(1+1)} analyzed the time complexity in the case of (1+1) EA for pseudo-Boolean functions such as Onemax and LeadingOnes, and subsequently demonstrated that the average running time of the (1+1) EA for linear pseudo-Boolean functions is $\Theta(n\ln n)$. Witt \cite{witt2013(1+1)} applied multiplicative drift analysis to obtain a tight bound of the expected running time in the case of $(1+1)$ EA for linear functions.

For the combinatorial optimization problem, Oliveto et al.\cite{oliveto2009(1+1)} analyzed the running time in the case of (1+1) EA for some specific instances of the vertex cover problem. Zhou et al.\cite{lai2013(1+1),zhou2014(1+1),zhou2015(1+1)} conducted approximation performance analyses in the case of (1+1) EA for instances of the minimum label spanning tree problem, multiprocessor scheduling problem, and maximum cut problem. Neumann \cite{neumann2018running-time} proved the time complexity upper bound in the case of (1+1) EA for the knapsack problem with favorably correlated weights by using multiplicative drift.
\subsection{$(1+\lambda)$ EA}
In the (1+$\lambda$) EA, the size of the parent population is 1 while the size of the offspring population is $\lambda$.

For the simple pseudo-Boolean problem, Jansen et al.\cite{jansen2005(1+lambda)} investigated the impact of the size of the offspring population on the performance of EAs by analyzing the first hitting time of the $(1+\lambda)$ EA on functions such as OneMax and LeadingOnes. Gie{\ss}en and Witt \cite{giessen2017(1+lambda)} studied the expected running time in the case of the $(1+\lambda)$ EA for Onemax with mutation probability $\frac{c}{n}$ where $c>0$ is any constant. Doerr et al. \cite{doerr2013(1+lambda)} analyze the running time in the case of $(1+\lambda)$ EA for royal road function.

For the combinatorial optimization problem, Oliveto \cite{oliveto2008(1+lambda)} presented a worst-case approximation running-time analysis in the case of $(1+\lambda)$ RLS and $(1+\lambda)$ EA for the vertex cover problem.
\subsection{$(\mu+1)$ EA}
In the $(\mu+1)$ EA, the size of the parent population is $\mu$ while the size of the offspring population is 1.

For the simple pseudo-Boolean problem, Witt \cite{witt2006(mu+1)} first analyzed the running time of the $(\mu+1)$ EA for pseudo-Boolean functions and obtained the time complexity upper bounds in the case of $(\mu+1)$ EA for the Onemax, LeadingOnes, and SPC functions. Zheng et al.\cite{zheng2021(mu+1)} analyzed the time complexity in the case of $(\mu+1)$ EA for a variant of OneMax with the time-linkage property.

For the combinatorial optimization problem, Theile \cite{theile2009(mu+1)} proved that the $(\mu+1)$ EA can find an optimal solution for TSP problem in non-trivial running time $O(n^{3}\cdot2^{n})$ when $\mu\le O(n\cdot2^{n})$. Nallaperuma et al.\cite{nallaperuma2013(mu+1)} analyzed the time complexity of $(\mu+1)$ EA for a class of TSP instances. Sutton \cite{sutton2018(mu+1)} proved the time complexity of a multi-start $(\mu+1)$ EA for any instances of the closest string problem.
\subsection{$(\mu+\lambda)$ EA}
In the $(\mu+\lambda)$ EA, the size of the parent population is $\mu$ while the size of the offspring population is $\lambda$.

For the simple pseudo-Boolean problem, Chen et al.\cite{chen2009new} analyzed the time complexity in the case of $(N+N)$ EA for Onemax and LeadingOnes. Qian et al.\cite{qian2016(mu+lambda)} prove the time complexity lower bound in the case of $(\mu+\lambda)$ EA for the class of pseudo-Boolean functions with a unique global optimum. Doerr \cite{doerr2020(mu+lambda)} performed a running-time analysis in the case of non-elitist $(\mu,\lambda)$ EA for the jump function. Lehre and Nguyen \cite{lehre2019(mu+lambda)} showed the expected running time in the case of non-elitist $(\mu,\lambda)$ EA for a benchmark problem Deceptive Leading Blocks.

For the combinatorial optimization problem, Sutton et al.\cite{sutton2014tsp} carried out a parameterized running-time analysis in the case of $(\mu+\lambda)$ EA for the planar Euclidean TSP. Xia et al.\cite{2019tsp} introduced an approximation running-time analysis for the TSP(1,2) problem.

The running-time analysis of population-based EAs for combinatorial optimization problems is still in its early stage compared to pseudo-Boolean problems. Our theoretical investigation of $(\mu+\lambda)$ EA for combinatorial optimization problems establishes an essential foundation for future research of analyzing EAs in practice.
\section{Multiple-Gain Model for Expected First Hitting Time Analysis of a Stochastic Process}
This section begins with an overview of fundamental concepts in single-objective combinatorial optimization, followed by the average gain model. 

\subsection{Multiple-Gain Model}
In this paper, we consider the following optimization problem\cite{back1997evolutionary}:
\begin{center}
	$min\quad f(x)\in S$\\
	$s. t. \quad x\in \left\{0, 1\right\}^{n}$
\end{center}
where $f(x)$ maps $\left\{0, 1\right\}^{n}$ to the range of fitness value $S$, with $n$ representing the encoding length.

Let $r_{i}\in S$ denote a fitness value, and $r_{m}\in S$ denote the biggest fitness value where $0\le i \le m$. For convenience, we presume $ S=\left\{r_{0}, r_{1}, \dots, r_{m}\right\}$ where $r_{0}<r_{1}<\dots<r_{m}$. In addition, we assume that for $\forall r_{i}\in S$, $0<\alpha\le r_{i}-r_{i-1}\le \beta$. The value of $\alpha$ represents the minimum difference between the fitness values of adjacent individuals in the set $S$, while the value of $\beta$ represents the maximum difference between the fitness values of adjacent individuals.

Let $x_{opt}\in \left\{0, 1\right\}^{n}$ denote a global optimum where $f(x_{opt})=r_{0}$, and $X_{t}= \left\{x_{1}, \dots,  x_{\mu}\right\}$ denote the population in the $t$-th generation with $\mu$ individuals $x_{i}\in \left\{0, 1\right\}^{n}$, $1\le i\le \mu$. Let $x_{t}=\left\{x_{t,1},x_{t,2},\dots,x_{t,n} \right\}$ denote any best individual in the population $X_{t}$ where for $\forall x\in X_{t}$, $f(x_{t})\ge f(x)$. Let $b_{t}$ denote the number of best individuals in the population $X_{t}$ where $1\le b_{t}\le \mu$.

The optimization process of EAs is characterized by randomness. Within the same parent population, it is possible to generate different offspring populations. Therefore, the optimization process can be regarded as a stochastic process. Let $(\Omega, F, P)$ be a probability space, and let $\left \{ Y_{t} \right \} _{t=0}^{\infty}$ represent a non-negative stochastic process defined on this space. Let $F_{t}=\sigma(Y_{0}, Y_{1}, \dots, Y_{t})$ denote the natural filtration of $F$.

In the following, we give the specific definitions of supermartingale and stopping time\cite{durrett2019}. 

\textbf{\emph{Definition 1: }}{Let $\left \{ Y_{t} \right \}_{t=0}^{\infty}$ and $\left \{ Z_{t} \right \}_{t=0}^{\infty}$ be two stochastic processes. $\left \{ Y_{t} \right \}_{t=0}^{\infty}$ is a supermartingale relating to $\left \{ Z_{t} \right \}_{t=0}^{\infty}$ if $Y_{t}$ depends on $Z_{0},Z_{1},\dots,Z_{t}$, $E(max\left\{0, -Y_{t}\right\})<\infty$, $E(Y_{t+1}|Z_{0}, Z_{1}, \dots, Z_{t})\le Y_{t}$ for all $ t\ge 0$. $\blacksquare$}

\textbf{\emph{Definition 2: }}{Let $\left \{ Y_{t} \right \}_{t=0}^{\infty}$ be a stochastic process and $T\in\mathbb{N}_0$ be a random variable. $T$ is a stopping time relating to $\left \{ Y_{t} \right \}_{t=0}^{\infty}$ if $\left\{T\le n\right\}\in F_{t}$ for all $n=0, 1, 2, \dots$. $\blacksquare$}

The iteration at which the global optimum is first found is called the first hitting time (FHT)\cite{he2016average}. The definition of FHT is given below. 

\textbf{\emph{Definition 3: }}{Let $\left \{ Y_{t} \right \}_{t=0}^{\infty}$ denote a stochastic process, where $Y_{t}\ge 0$ holds for all $t\ge 0$. The first hitting time of EAs is defined as $T_{\varepsilon}=min\left\{t\ge 0: Y_{t}\le \varepsilon\right\}$ where $\varepsilon$ represents the target accuracy of EAs. }
Specifically, $T_{0}=min\left\{t\ge 0:Y_{t}=0\right\}$ is a stopping time relating to $\left \{ Y_{t} \right \}_{t=0}^{\infty}$ \cite{yushan2016first}. $\blacksquare$

Moreover, the expected first hitting time (EFHT) of EAs is denoted by $E(T_{\varepsilon})$, which is the expected value of FHT. The EFHT represents the iteration number on average needed for EAs to first attend a global optimum\cite{yu2008new}. 
The average gain represents the difference between the fitness value of the best individual $x_{t}$ in the parent population and the fitness value of the best individual $x_{t+1}$ in the offspring population. The formal definition of average gain is presented below\cite{yushan2016first}.

\textbf{\emph{Definition 4: }}{The gain at generation $t$ is}
\begin{align*}
	g_{t}=f(x_{t})-f(x_{t+1}). 
\end{align*}
Let $H_{t}=\sigma(f(x_{0}), f(x_{1}), \dots, f(x_{t}))$. The average gain at generation $t$ is 
\begin{align*}
	E(g_{t}|H_{t})=E(f(x_{t})-f(x_{t+1})|H_{t}).\quad \hfill{\blacksquare} 
\end{align*}

Similar to quality gain, the average gain indicates the expected progress of EAs within one iteration\cite{akimoto2017quality}. A greater average gain implies a quicker convergence towards the global optimum, thereby improving the efficiency of each iteration in the optimization process.

Based on the average gain model, we propose the multiple-gain model for analyzing the EFHT of evolutionary combinatorial optimization. The definitions of multiple-gain and expected multiple-gain are presented in Definition 5.

\textbf{\emph{Definition 5: }}{For a given $k$, the multiple-gain at generation $t$ is}
\begin{align*}
	f(x_{t})-f(x_{t+k}). 
\end{align*}
Let $H_{t}=\sigma(f(x_{0}), f(x_{1}), \dots, f(x_{t}))$. For a given $k$, the expected multiple-gain at generation $t$ is 
\begin{align*}
	G(t,k)=E(f(x_{t})-f(x_{t+k})|H_{t}).\quad \hfill{\blacksquare}  
\end{align*}

The average gain is a special case of the expected multiple-gain model when $k=1$.

\subsection{Estimating worst-case upper bounds of EFHT for a stochastic process}
Based on the expected multiple-gain, the worst-case upper bound of EFHT can be estimated using Theorem 1. The following lemma, which was proved in \cite{yushan2016first}, will be used in the proof of Theorem 1.

\textbf{\emph{Lemma 1: }}{Let $\left \{ S_{t} \right \} _{t=0}^{\infty}$ be a supermartingale with respect to
	$\left \{ Y_{t} \right \}_{t=0}^{\infty}$ and $T$ be a stopping time relating to $\left \{ Y_{t} \right \} _{t=0}^{\infty}$,}
\begin{align}
	E(S_{T}|F_{0})\le S_{0}. 
\end{align}

\textbf{\emph{Theorem 1: }}{Let $\left\{f(x_{t})\right\}_{t=0}^{\infty}$ denote a stochastic process, where $f(x_{t})\ge 0$ holds for all $t\ge 0$. Assume that $f(x_{t})=r_{i}$, where $0<i\le m$, $t<T_{0}$. If there exists $k\in \mathbb{N}^+$ such that $G(t,k)=E(f(x_{t})-f(x_{t+k})|H_{t})\ge r_{i}-r_{i-1}\ge \alpha>0$ holds for all $t<T_{0}$, }
\begin{align}
	E(T_{0}|f(x_{0}))\le \frac{kf(x_{0})}{\alpha}. 
\end{align}

\emph{Proof:} Define $Q_{t}=f(x_{kt})+\frac{t}{k}\alpha, t=0, 1, 2, \dots$. Let $H_{kt}=\sigma (f(x_{0}), f(x_{k}), \dots, f(x_{kt}))$, for $\forall kt<T_{0}$,
\begin{align*}
	E(Q_{t+1}-Q_{t}|H_{kt})&=E(f(x_{kt+k})-f(x_{kt})+\frac{\alpha}{k} |H_{kt})\\
	&=E(f(x_{kt+k})-f(x_{kt})|H_{kt})+\frac{\alpha}{k}\\
	&=-G(kt,k)+\frac{\alpha}{k}\\
	&\le -\alpha+\frac{\alpha}{k}\\
	&< 0. 
\end{align*}

This means that
\begin{align*}
	E(Q_{t+1}|H_{kt})&\le E(Q_{t}|H_{kt})\\
	&=Q_{t}. 
\end{align*}

\begin{spacing}{1}
	By Definition 1, $\left\{Q_{t}\right\}_{t=0}^{\infty}$ is a supermartingale relating to $\left\{f(x_{kt})\right\}_{t=0}^{\infty}$. Consider $T_{0}$ is a stopping time relating to stochastic process$\left \{ f(x_{kt}) \right \} _{t=0}^{\infty}$. By Lemma 1, we have 
	\begin{center}
		$E(Q_{T_{0}}|H_{0})\le Q_{0}=f(x_{0})$. 
	\end{center}
\end{spacing}

Therefore,
\begin{align*}
	f(x_{0})
	&\ge E(Q_{T_{0}}|H_{0})\\
	&=E(f(x_{kT_{0}})+\frac{T_{0}}{k}\alpha|H_{0})\\
	&=E(f(x_{kT_{0}})|H_{0})+\frac{\alpha E(T_{0}|H_{0})}{k}\\
	&\ge \frac{\alpha E(T_{0}|H_{0})}{k}.
\end{align*}

\begin{spacing}{1}
	This means that $\displaystyle E(T_{0}|H_{0})=E(T_{0}|f(x_{0}))\le \frac{kf(x_{0})}{\alpha}. $ $\hfill{\blacksquare}$
\end{spacing}

\begin{spacing}{1}
	The conclusion of Theorem 1 in this paper differs from that of Theorem 1 in \cite{yushan2016first}, as the former is based on expected multiple-gain, while the latter relies on average gain. Moreover, while the lower bound of average gain needs to be a constant in \cite{yushan2016first}, the lower bound of the expected multiple-gain depends on the range of fitness value $S=\left\{r_{0}, r_{1}, \dots, r_{m}\right\}$ in this paper.

	Theorem 1 describes a scenario where the fitness value changes only after every $k$ iteration, and the magnitude of the change is the smallest possible increment $\alpha$. Notably, the value of $k$ is related to the longest duration during which the gain remains zero in the execution process of EAs. Therefore, the conclusion of Theorem 1 represents the worst-case upper bound of EFHT. Furthermore, a high value of $k$ can lead to a less accurate worst-case upper bound on $E(T_{0})$. To improve the accuracy of the worst-case upper bound on $E(T_{0})$, it is necessary to obtain the lower bound of $k$. The definition of $k_{low}$ is presented below.
\end{spacing}

\textbf{\emph{Definition 6: }}{Assume that $f(x_{t})=r_{i}$, where $0<i\le m$, $t<T_{0}$. $k_{low}\in\mathbb{N}^+$ is the lower bound of $k$ if $G(t,k)\ge r_{i}-r_{i-1}$ holds for all $t<T_{0}$, $k\ge k_{low}$. $\blacksquare$ }

By Theorem 1 and Definition 6, we can derive the following Corollary 1.

\textbf{\emph{Corollary 1: }}{Let $\left\{f(x_{t})\right\}_{t=0}^{\infty}$ be a stochastic process, where $f(x_{t})\ge 0$ holds for all $t\ge 0$. Assume that $f(x_{t})=r_{i}$, where $0<i\le m$, $t<T_{0}$. If there exists $k\in \mathbb{N}^+$ such that $G(t,k)=E(f(x_{t})-f(x_{t+k})|H_{t})\ge r_{i}-r_{i-1}\ge \alpha>0$ holds for all $t<T_{0}$, }
\begin{align}
	E(T_{0}|f(x_{0}))\le \frac{k_{low}f(x_{0})}{\alpha}. 
\end{align}

\emph{Proof:} By Definition 6, we have
\begin{align*}
	E(f(x_{t})-f(x_{t+k_{low}})|H_{t})\ge r_{i}-r_{i-1}\ge \alpha
\end{align*}

According to Theorem 1, we have
\begin{align*}
	E(T_{0}|f(x_{0}))\le \frac{k_{low}f(x_{0})}{\alpha}.\quad \hfill{\blacksquare}  
\end{align*}

Since $k_{low}$ is the minimum value of $k$ that satisfies $G(t,k)=E(f(x_{t})-f(x_{t+k})|H_{t})\ge r_{i}-r_{i-1}$, the upper bound provided by Corollary 1 is in fact the least upper bound in Theorem 1. Therefore, the precise estimation of $k_{low}$ is crucial to estimate the worst-case upper bound of EFHT. Corollary 1 will be utilized in Sections IV to estimate the worst-case upper bounds of EFHT.

\subsection{Estimating average-case upper bounds of EFHT for a stochastic process}
 If we do not restrict the value of $k$, the average-case upper bound of EFHT can be estimated using Theorem 2.

\textbf{\emph{Theorem 2: }}{Let $\left \{ f(x_{t}) \right \}_{t=0}^{\infty}$ denote a stochastic process, where $f(x_{t})\ge 0$ holds for all $t\ge 0$. If there exists a monotonically non-decreasing function $h(r):\left\{0,1\right\}^{n}\rightarrow R^{+}$ and $k\in \mathbb{N}^+$ such that $G(t,k)=E(f(x_{t})-f(x_{t+k})|H_{t})\ge h(f(x_{t}))$ holds for all $t<T_{0}$, }
\begin{align}
	E(T_{0}|f(x_{0})=r_{L})\le k\cdot \sum_{i=1}^{L}\frac{r_{i}-r_{i-1}}{h(r_{i})}. 
\end{align}

\emph{Proof:} Let $f(x_{t})=r_{p}$, $f(x_{t+k})=r_{q}$, where $0\le q<p<L$. 
Let $g(f(x_{t}))=\begin{cases}
	0,f(x_{t})=0\\
	\displaystyle \sum_{i=1}^{p}\frac{r_{i}-r_{i-1}}{h(r_{i})},f(x_{t})>0
\end{cases}$, for $\forall t<T_{0}$,

(1) When $t<T_{0}$, $t+k<T_{0}$,
\begin{align*}
	g(f(x_{t}))-g(f(x_{t+k}))&=\sum_{i=1}^{p}\frac{r_{i}-r_{i-1}}{h(r_{i})}-\sum_{i=1}^{q}\frac{r_{i}-r_{i-1}}{h(r_{i})}\\
	&=\sum_{i=q+1}^{p}\frac{r_{i}-r_{i-1}}{h(r_{i})}.
\end{align*}

Therefore,
\begin{align*}
	E\left[g(f(x_{t}))-g(f(x_{t+k}))|H_{t}\right]
	&=E(\sum_{i=q+1}^{p}\frac{r_{i}-r_{i-1}}{h(r_{i})}|H_{t})\\
	&\ge E(\frac{r_{p}-r_{q}}{h(r_{p})}|H_{t})\\
	&=E(\frac{f(x_{t})-f(x_{t+k})}{h(f(x_{t}))}|H_{t})\\
	&=\frac{G(t,k)}{h(f(x_{t}))}. 
\end{align*}

According to the assumption in Theorem 2, there exists a monotonically non-decreasing function $h(r)$ that satisfies $G(t,k)\ge h(f(x_{t}))$. Therefore,
\begin{align*}
	E\left[g(f(x_{t}))-g(f(x_{t+k}))|H_{t}\right]\ge 1. 
\end{align*}

(2) When $t<T_{0}$, $t+k\ge T_{0}$,
\begin{align*}
	g(f(x_{t}))-g(f(x_{t+k}))&=\sum_{i=1}^{p}\frac{r_{i}-r_{i-1}}{h(r_{i})}.
\end{align*}

Therefore,
\begin{align*}
	E\left[g(f(x_{t}))-g(f(x_{t+k}))|H_{t}\right]
	&=E(\sum_{i=1}^{p}\frac{r_{i}-r_{i-1}}{h(r_{i})}|H_{t})\\
	&\ge E(\sum_{i=q+1}^{p}\frac{r_{i}-r_{i-1}}{h(r_{i})}|H_{t})\\
	&\ge 1
\end{align*}

Note that
\begin{align*}
    T_{0}&=min\left\{t\ge 0:f(x_{t})=0\right\}\\
    &=min\left\{t\ge 0:g(f(x_{t}))=0\right\}\stackrel{\wedge}{=}T_{0}^{g}
\end{align*}

According to Theorem 1, we have
\begin{align*}
	E(T_{0}|f(x_{0}))=E\left[T_{0}^{g}|g(f(x_{0}))\right]
	&\le \frac{k\cdot g(f(x_{0}))}{1}.
\end{align*}

\begin{spacing}{1}
	This means $E(T_{0}|f(x_{0}))\le k\cdot \sum_{i=1}^{L}\frac{r_{i}-r_{i-1}}{h(r_{i})}$. $\hfill\blacksquare$
\end{spacing}

\begin{spacing}{1}
	The conclusion of Theorem 2 in this paper differs from that of Theorem 2 in \cite{yushan2016first}, as the former is established based on expected multiple-gain, while the latter is built upon average gain. Furthermore, the lower bound of average gain must be a monotonically increasing and integrable function in Theorem 2 of \cite{yushan2016first}, while the lower bound of expected multiple-gain is a monotonically non-decreasing function in Theorem 2 of this paper. 

Formula (4) is determined by the initial solution $x_{0}$, the lower bound $h(r_{p})$ of $G(t,k)$, the parameter $k$ of $G(t,k)$ and the range of fitness value $S$. In addition, $G(t,k)\ge h(r_{p})$ indicates that the fitness value may not necessarily decrease after $k$ iterations, because $h(r_{p})$ is not always greater than $r_{p}-r_{p-1}$. Therefore, $k$ can be any positive integer if there exists a function $h(r_{p})$ that satisfies $G(t,k)\ge h(r_{p})$. This means that the conclusion of Theorem 2 represents the average-case upper bound of EFHT. Theorem 2 will be utilized in Section IV to analyze the average-case upper bounds of EFHT. 
\end{spacing}





\section{Expected First Hitting Time Analysis of EAs Based on Multiple-gain Estimation}
    This section will present a detailed EFHT analysis of evolutionary combinatorial optimization instances using the multiple-gain model. 
    
    \subsection{Procedure of Multiple-gain Estimation for  Expected First Hitting Time Analysis of EAs }
    Fig. 1 illustrates the process of EFHT analysis using the multiple-gain model. We can use the multiple-gain model to estimate the worst-case upper bounds of EFHT by Formula (3). According to the law of large numbers, the larger the sample size, the higher the probability that the arithmetic mean will approach the expected value\cite{durrett2019}. Therefore, we can estimate $k_{low}$ by taking the average of each longest duration during which the gain remains zero obtained from repeated runs of the algorithm. Let $N$ denote the number of independent repeated runs of the algorithm. Let $k_{i}$ denote the longest interval with zero gain during the $i$-th run. The $k_{low}$ can be estimated as follows
    \begin{align}
        k_{low} = \frac{\sum_{i=1}^{N}k_{i}}{N}
    \end{align}
    To utilize Formula (3), we still need the value of $\alpha$, which represents the minimum difference between the fitness values of adjacent individuals in the set $S$. The value of $\alpha$ can be estimated through experimental data. Specifically, we can record all data of gain that are not equal to zero in the experiment. There will be at least one gain equal to $\alpha$ with a sufficiently large number of experiments. Let $G$ be the set of all non-zero gains obtained from $N$ repeated executions of the algorithm. Therefore, the $\alpha$ can be estimated as follows
    \begin{align}
        \alpha = \min_{g \in G} g
    \end{align}
    Once the estimated $k_{low}$ and $\alpha$ are obtained, the worst-case upper bound of EFHT can be estimated using Formula (3).

The multiple-gain model estimates the average-case upper bounds of EFHT by Formula (4). Before deriving the average-case upper bound of EFHT, it is necessary to select the parameter $k$ for expected multiple-gain, where the range of $k$ depends on the estimation of $k_{low}$ obtained from experiments. 

\begin{figure*}
	\centering
	\includegraphics[width=0.85\textwidth]{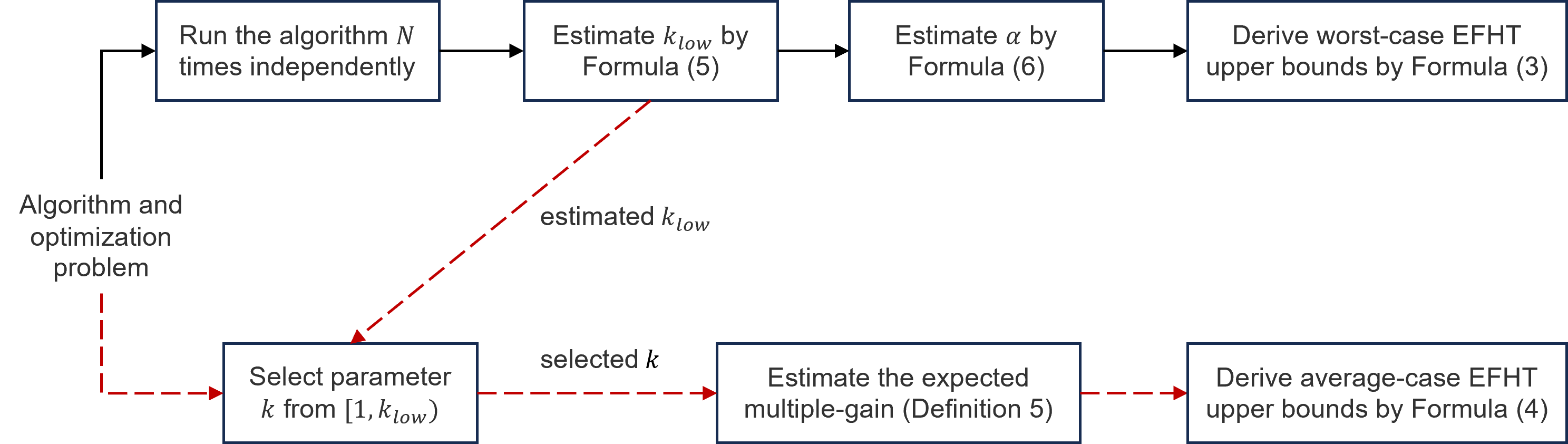}
	\caption{The EFHT analysis process using the multiple-gain model}
\end{figure*}


In the following three subsections, we utilize Theorem 2 to analyze the average-case upper bounds of EFHT for $(\mu+\lambda)$ EA on three combinatorial optimization problems. The
expected multiple-gain varies with different $k\in \left[1,k_{low}\right)$,
leading to different average-case upper bounds of EFHT.
Regardless of the value of $k$, we need to compute the expected multiple-gain according to Definition 5 and then derive the average-case upper bounds of EFHT through Formula (4). Therefore, we set $k=1$ to compute the expected multiple-gain for simplifying the analysis procedure. Although setting $k$ to 1 is a relatively simple approach, we still derive tighter average-case upper bounds of EFHT. In addition, the state of individual $x_{t+1}$ depends only on the individual $x_{t}$ for all the problems discussed in the following subsections. Therefore, $\left\{f(x_{t})\right\}_{t=0}^{\infty}$ can be modeled as a Markov chain. Consequently, the expected multiple-gain can be rewritten as
\begin{align*}
	G(t, 1)=E(f(x_{t})-f(x_{t+1})|f(x_{t})). 
\end{align*} 
Moreover, we conduct analyses to determine the theoretical $k_{low}$ that satisfies Theorem 1 for these evolutionary combinatorial optimization instances. These analyses validate the applicability of the multiple-gain model for different evolutionary combinatorial optimization instances.

 \subsection{Case Study I:  Expected First Hitting Time Analysis of ($\mu +\lambda$) EA for the Knapsack Problem with Favorably Correlated Weights}

The knapsack problem is one of the most important optimization problems. The objective of the knapsack problem is to maximize a linear profit function while satisfying a linear weight constraint \cite{neumann2019running-time}. For $x\in \left\{0, 1\right\}^{n}$, the knapsack problem is typically formulated as follows:
\begin{align*}
	max\quad f(x)=\sum_{i=1}^{n}v_{i}x_{i}\\
	s. t \quad B(x)=\sum_{i=1}^{n}w_{i}x_{i}\le K
\end{align*}
where $w_{i}>0$ denotes the weight of item $i$, $v_{i}>0$ denotes the value of item $i$, and $K$ represents the capacity of knapsack. If $x_i$ is set to 1, this indicates that the item $i$ is to be included in the knapsack. Conversely, if $x_i$ is set to 0, the item $i$ will be excluded from the knapsack. 

\begin{spacing}{1}
	We consider an instance of the knapsack problem where for two different items $i$ and $j$, the conditions $v_{i}\ge v_{j}$ and $w_{i}\le w_{j}$ hold for all $i<j$. This instance represents a generalization of the linear function optimization problem with a uniform constraint, as previously explored in\cite{friedrich2017analysis}. Neumann et al.\cite{neumann2018running-time} proved that the time complexity upper bound of (1 + 1) EA for this instance is $O(n^{2}(\ln n+p_{max}))$, where $p_{max}$ is the largest profit of given items. Theorem 3 gives a more general upper bound of EFHT of ($\mu+\lambda)$ EA for this instance. Detailed operations of ($\mu+\lambda$) EA are described as Algorithm 1 \cite{qian2016(mu+lambda)}. 
\end{spacing}

\begin{algorithm}
	\caption{($\mu+\lambda$) EA for the knapsack problem}
	\begin{algorithmic}[1]
		\REQUIRE {Encoding length $n$.}
		\ENSURE {The global optimum $x_{opt}$.}
		\STATE {Initialization: set generation $t$=0, generate a population 
			$X_{0}= \left\{x_{1}, \dots,  x_{\mu}\right\}$ with $\mu$ individuals $x_{i}\in \left\{0, 1\right\}^{n}$, $1\le i\le \mu$, uniformly at random.}
		\WHILE{a global optimum has not been attended}
            \STATE ${X}'=X_{t}$
		\FOR{$i=1$ to $\lambda$}
            \STATE choose a $x$ from $X_{t}$ uniformly at random
            \STATE $x_{temp}=x$
		\STATE ${x}' \leftarrow$ flip each bit of $x$ independently with probability $\frac{1}{n}$
            \IF{$f({x}')>f(x_{t})$ and $x_{temp}\ne x_{t}$} 
            \STATE ${x}' \leftarrow x_{temp}$
            \ENDIF
            \STATE ${X}'\leftarrow {X}'\cup\left\{{x}'\right\}$
		\ENDFOR
		\STATE  delete the $\lambda$ individuals with the lowest fitness value of ${X}'$ uniformly at random
            \STATE $X_{t+1} \leftarrow$ ${X}'$
		\STATE $t \leftarrow t+1$
		\ENDWHILE
            \STATE $x_{opt} \leftarrow$ choose a $x$ from $X_{t}$ with the maximum fitness value 
		\RETURN $x_{opt}$
	\end{algorithmic}
\end{algorithm}

For analysis, Algorithm 1 is configured to reject any offspring solutions ${x}'$ generated by the non-best individuals if $f({x}')>f(x_{t})$ (lines 8-10). This restriction cause Algorithm 1 to miss opportunities for progress, leading to increased running time. Under this rule, $x_{t+1}$ must be created by $x_{t}$ if $f(x_{t+1})>f(x_{t})$.

Let $d_{min}=min_{v_{i}\ne v_{j}}\left\{|v_{i}-v_{j}|\right\}$, $v_{min}=min\left\{v_{i}|1\le i\le n\right\}$. $d_{min}$ and $v_{min}$ will be utilized in the subsequent theorems.

\textbf{\emph{Theorem 3: }}{Suppose $x_{opt}=\left \{\underset{q}{\underbrace{1,1,\dots,1}},0,0,\dots,0 \right \}$. Let $Y_{t}=f(x_{opt})-f(x_{t})$ and $T_{0}=min\left\{t\ge 0:Y_{t}=0\right\}$, the average-case EFHT of Algorithm 1 for the knapsack problem with favorably
	correlated weights satisfies}
\begin{align}
	E(T_{0}|Y_{0})&\le \frac{(Y_{0}-r_{0})}{(1-e^{-\frac{\lambda (p_{1}+n(1-p_{1})}{\mu n^{2}e}})(p_{1}d_{min}+p_{2}v_{min})}
\end{align}
where $p_{1}=1-(\sum_{i=0}^{q-1}C_{q}^{i})/N$, $p_{2}=(\sum_{i=0}^{q-1}C_{q}^{i})/N$ and $N$ denotes the number of feasible solutions.

\emph{Proof:} Note that there is at least one 0-bit in the first $q$ bits for any $x_{t}$ where $t<T_{0}$. Therefore, the proof will be discussed in two cases. One case examines the scenario with at least one 1-bit in the last $n-q$ bits of $x_{t}$. The other case addresses the scenario where the last $n-q$ bits of $x_{t}$ are all 0-bits.


\textbf{Case 1}: There is at least one item $i$ such that $x_{t, i}=1$, $t<T_{0}$ and $q<i\le n$. Since $t<T_{0}$, there must exist an item $j$ such that $1\le j \le q, x_{t, j}=0$. Therefore, $x_{t}$ can be optimized by flipping $x_{t,i}$ and $x_{t,j}$. The probability $p_{i,j}$ that flipping $x_{t,i}$ and $x_{t,j}$ satisfies
\begin{align*}
    p_{i,j}=(\frac{1}{n})^{2}(1-\frac{1}{n})^{n-2}\ge \frac{1}{n^{2}e}
\end{align*}

Therefore,
\begin{align*}
    E(Y_{t}-Y_{t+1}|Case 1)&\ge \frac{1}{n^{2}e}(v_{i}-v_{j})
    \ge \frac{1}{n^{2}e}d_{min}
\end{align*}


\textbf{Case 2}: There does not exist any items $i$ such that $x_{t, i}=1$, $t<T_{0}$, and $q<i\le n$. Since $t<T_{0}$, there must exist an item $j$ such that $1\le j \le q$ and $x_{t, j}=0$. Therefore, $x_{t}$ can be optimized by flipping $x_{t,j}$. The probability $p_{j}$ that flipping $x_{t,j}$ satisfies
\begin{align*}
    p_{j}=\frac{1}{n}(1-\frac{1}{n})^{n-1}\ge \frac{1}{ne}
\end{align*}

Therefore,
\begin{align*}
    E(Y_{t}-Y_{t+1}|Case 2)&\ge \frac{1}{ne}v_{i}
    \ge \frac{1}{ne}v_{min}
\end{align*}


Let $p_{1}$ denote the probability of Case 1 and $p_{2}$ denote the probability of Case 2. Note that Case 2 represents the situation where the last $n-q$ bits of $x_{t}$ are all 0-bits. Furthermore, there are at most $q-1$ bits in the first $q$ bits of $x_{t}$ are 1-bits. Let $N$ denote the number of feasible solutions. Therefore, $p_{2}=(\sum_{i=0}^{q-1}C_{q}^{i})/N$, $p_{1}=1-p_{2}=1-(\sum_{i=0}^{q-1}C_{q}^{i})/N$. The probability $p_{3}$ of creating an offspring individual $x$ by $x_{t}$ where $f(x)>f(x_{t})$ satisfies
\begin{align*}
    p_{3}\ge\frac{b_{t}}{\mu}\left[p_{1}\frac{1}{n^{2}e}+p_{2}\frac{1}{ne}\right]=\frac{b_{t}}{\mu}\frac{p_{1}+n(1-p_{1})}{n^{2}e}
\end{align*}


The probability $p_{4}$ that this event occurs at least once during $\lambda$ generations is 
\begin{align*}
    p_{4}&=1-(1-\frac{b_{t}(p_{1}+n(1-p_{1})}{\mu n^{2}e})^{\lambda}
\end{align*}

When $0<a\le b$ and $n>0$, we have $a^{n}\le b^{n}$. Furthermore, $(1+x)\le e^{ x}, \forall x\in R$. Therefore, when $x>-1$ and $\lambda>0$, we have,
\begin{align*}
	(1+x)^{\lambda}\le e^{\lambda x}
\end{align*}

As $-\frac{b_{t}(p_{1}+n(1-p_{1})}{\mu n^{2}e}>-1$, we have
\begin{align*}
    p_{4}
    &\ge 1-e^{-\frac{\lambda b_{t}(p_{1}+n(1-p_{1})}{\mu n^{2}e} }\\
    &\ge 1-e^{-\frac{\lambda(p_{1}+n(1-p_{1})}{\mu n^{2}e} }
\end{align*}

Therefore,
\begin{align*}
    G(t,1)&=E(Y_{t}-Y_{t+1}|Y_{t})\\
    &\ge p_{4}(p_{1}d_{min}+p_{2}v_{min})\\
    &\ge (1-e^{-\frac{\lambda (p_{1}+n(1-p_{1})}{\mu n^{2}e}})(p_{1}d_{min}+p_{2}v_{min})
\end{align*}








Let $h_{1}(r)=(1-e^{-\frac{\lambda (p_{1}+n(1-p_{1})}{\mu n^{2}e}})(p_{1}d_{min}+p_{2}v_{min})$. Note that $h_{1}(r)$ is a constant that can be considered a monotonically non-decreasing function. By Theorem 2, we have 
\begin{align*}
	E(T_{0}|Y_{0}=r_{L})&\le \sum_{i=1}^{L}\frac{r_{i}-r_{i-1}}{(1-e^{-\frac{\lambda (p_{1}+n(1-p_{1})}{\mu n^{2}e}})(p_{1}d_{min}+p_{2}v_{min})}\\
	&= \frac{(r_{L}-r_{0})}{(1-e^{-\frac{\lambda (p_{1}+n(1-p_{1})}{\mu n^{2}e}})(p_{1}d_{min}+p_{2}v_{min})}
\end{align*}
 $\hfill{\blacksquare}$

Theorem 3 presents a more general result than that found in \cite{neumann2018running-time}. We will show the average running time of Algorithm 1 in Section V to validate the correctness of Theorem 3. Moreover, the conclusion of Theorem 3 can be discussed in two distinct cases, leading to the following Corollary 2.

\textbf{\emph{Corollary 2: }}{Let $Y_{t}=f(x_{opt})-f(x_{t})$ and $T_{0}=min\left\{t\ge 0:Y_{t}=0\right\}$, the EFHT of Algorithm 1 for the knapsack problem with favorably
	correlated weights satisfies}
\begin{align}
	E(T_{0}|Y_{0}=r_{L})\in 
	\begin{cases}
		O(\frac{\mu n(r_{L}-r_{0})}{\lambda v_{min}}),p_{1}=0
		\\
		O(\frac{\mu n^{2}(r_{L}-r_{0})}{\lambda d_{min}}),p_{1}=1
	\end{cases}
\end{align}
\emph{Proof:} By Theorem 3, we have
\begin{align*}
	E(T_{0}|Y_{0}=r_{L})
	&\le \frac{(r_{L}-r_{0})}{(1-e^{-\frac{\lambda (p_{1}+n(1-p_{1})}{\mu n^{2}e}})(p_{1}d_{min}+p_{2}v_{min})}
\end{align*}

(1) When $p_{1}=0$, $p_{2}=1$,
\begin{align*}
	E(T_{0}|Y_{0}=r_{L})&\le \frac{(r_{L}-r_{0})}{(1-e^{-\frac{\lambda n}{\mu n^{2}e}})v_{min}}\\
    &=\frac{(r_{L}-r_{0})}{(1-e^{-\frac{\lambda}{\mu ne}})v_{min}}
\end{align*}

When $n\rightarrow \infty$,
\begin{align*}
    \lim_{n \to \infty}1-e^{-\frac{\lambda}{\mu ne}}&=\lim_{n \to \infty}1-(1-\frac{\lambda}{\mu ne})\\
    &=\frac{\lambda}{\mu ne}
\end{align*}


This means $E(T_{0}|Y_{0}=r_{L})\in O(\frac{\mu n(r_{L}-r_{0})}{\lambda v_{min}})$.

(2) When $p_{1}=1$, $p_{2}=0$,
\begin{align*}
	E(T_{0}|Y_{0}=r_{L})&\le \frac{(r_{L}-r_{0})}{(1-e^{-\frac{\lambda}{\mu n^{2}e}})d_{min}}
\end{align*}

When $n\rightarrow \infty$,
\begin{align*}
    \lim_{n \to \infty}1-e^{-\frac{\lambda}{\mu n^{2}e}}&=\lim_{n \to \infty}1-(1-\frac{\lambda}{\mu n^{2}e})\\
    &=\frac{\lambda}{\mu n^{2}e}
\end{align*}


\begin{spacing}{1}
	This means $E(T_{0}|Y_{0}=r_{L})\in O(\frac{\mu n^{2}(r_{L}-r_{0})}{\lambda d_{min}})$. $\hfill{\blacksquare}$
\end{spacing}

\begin{spacing}{1}
Corollary 2 shows a new upper bound of EFHT $O(\frac{n^{2}(r_{L}-r_{0})}{d_{min}})$, which is tighter than the result $O(n^{2}(\ln n+p_{max}))$ in \cite{neumann2018running-time} by a factor of $\ln n$ when $\mu=1$, $\lambda=1$ and $p_{1}=1$. Furthermore, a better upper bound of EFHT $O(\frac{n(r_{L}-r_{0})}{v_{min}})$ which is tighter than the result in \cite{neumann2018running-time} by a factor of $n\ln n$ is also derived when $\mu=1$, $\lambda=1$ and $p_{1}=0$. 
\end{spacing}

By Corollary 1 and Theorem 3, we can obtain the worst-case EFHT of ($\mu+\lambda$) EA for the knapsack problem with favorably correlated weights.

\textbf{\emph{Theorem 4: }}{Suppose $x_{opt}=\left \{\underset{q}{\underbrace{1,1,\dots,1}},0,0,\dots,0 \right \}$. Let $Y_{t}=f(x_{opt})-f(x_{t})$ and $T_{0}=min\left\{t\ge 0:Y_{t}=0\right\}$, the worst-case EFHT of Algorithm 1 for the knapsack problem with favorably
	correlated weights satisfies}
\begin{align}
        E(T_{0}|Y_{0})&\le \frac{\beta(Y_{0}-r_{0})}{\alpha(1-e^{-\frac{\lambda (p_{1}+n(1-p_{1})}{\mu n^{2}e}})(p_{1}d_{min}+p_{2}v_{min})}
\end{align}
where $p_{1}=1-(\sum_{i=0}^{q-1}C_{q}^{i})/N$, $p_{2}=(\sum_{i=0}^{q-1}C_{q}^{i})/N$ and $N$ denotes the number of feasible solutions.

\emph{Proof:} By the proof of Theorem 3, we have $G(t, 1)\ge h_{1}(r_{p})$. Therefore, to achieve an expected multiple-gain of at least $r_{p}-r_{p-1}$ after $k$ iterations, $k$ must satisfy the condition $k\cdot h_{1}(r_{p})\ge r_{p}-r_{p-1}$. In addition, $\forall r_{i}\in S$, $0<\alpha\le r_{i}-r_{i-1}\le \beta$. By the inequality $k\cdot h_{1}(r_{p})\ge r_{p}-r_{p-1}$, we have 
\begin{align*}
	k\ge \frac{\beta}{h_{1}(r_{p})}
\end{align*}

Therefore, 
\begin{align}
    k_{low}=\frac{\beta}{(1-e^{-\frac{\lambda (p_{1}+n(1-p_{1})}{\mu n^{2}e}})(p_{1}d_{min}+p_{2}v_{min})}
\end{align}

By Corollary 1, we have
\begin{align}
    E(T_{0}|Y_{0}=r_{L})&\le \frac{k_{low}(r_{L}-r_{0})}{\alpha}\\
    &=\frac{\beta(Y_{0}-r_{0})}{\alpha(1-e^{-\frac{\lambda (p_{1}+n(1-p_{1})}{\mu n^{2}e}})(p_{1}d_{min}+p_{2}v_{min})} 
\end{align}
$\hfill{\blacksquare}$




We will estimate the $k_{low}$ and the worst-case upper bounds of EFHT through experimental data in Section V to validate the correctness of Theorem 4.

 \subsection{Case Study II: Expected First Hitting Time Analysis of ($\mu+\lambda$) EA for the $k$-MAX-SAT problem}

Existing running-time analysis studies of the SAT problem have focused on specific instances. For example, Buzdalov and Doerr et al. \cite{buzdalov2017running-time,doerr2015improved} analyzed the running time of EAs for solving random 3-CNF formulas. Sutton et al. \cite{sutton2012parameterized} conducted a running-time analysis of the 2-MAX-SAT problem. We will perform a running-time analysis of the general $k$-MAX-SAT problem to give a more comprehensive view of EAs on this problem.

A $k$-MAX-SAT problem instance is formulated as follows:
\begin{align*}
	l=\left\{(l_{1,1}\vee l_{1,2}\vee \dots l_{1,k}),\dots,(l_{s,1}\vee l_{s,2}\vee \dots l_{s,k})\right\}
\end{align*}
where $s$ denotes the number of clauses, the length of each clause is $k$, and $l_{i,j}$ denote a Boolean variable or its negation. The candidate solutions for the $k$-MAX-SAT instance are represented as binary strings of length $n$. Every bit of the binary string can be interpreted as the state of a Boolean variable $v_{i}$ (i.e., $x_{i}=1$ corresponds to $v_{i}=true$; $x_{i}=0$ corresponds to $v_{i}=false$). Let the function $f(x):\left\{0,1\right\}^{n}\rightarrow \left\{0,1,\dots,s\right\}$ count the clauses in $l$ that are satisfied under the assignment corresponding to $x$. Therefore, the
$k$-MAX-SAT problem is transformed into a pseudo-Boolean function optimization problem.

 Theorem 5 gives the upper bound of EFHT of ($\mu+\lambda$) EA with mutation probability $\frac{1}{2}$ for the $k$-MAX-SAT problem. Detailed operations of ($\mu+\lambda$) EA for the $k$-MAX-SAT problem are described as Algorithm 2 \cite{qian2016(mu+lambda)}. 
\begin{algorithm}
	\caption{($\mu+\lambda$) EA for the $k$-MAX-SAT problem}
	\begin{algorithmic}[1]
		\REQUIRE {Encoding length $n$.}
		\ENSURE {The global optimum $x_{opt}$.}
		\STATE {Initialization: set generation $t$=0, generate a population 
			$X_{0}= \left\{x_{1}, \dots,  x_{\mu}\right\}$ with $\mu$ individuals $x_{i}\in \left\{0, 1\right\}^{n}$, $1\le i\le \mu$, uniformly at random.}
		\WHILE{a global optimum has not been attended}
            \STATE ${X}'=X_{t}$
		\FOR{$i=1$ to $\lambda$}
            \STATE choose a $x$ from $X_{t}$ uniformly at random
		\STATE ${x}' \leftarrow$ flip each bit of $x$ independently with probability $\frac{1}{2}$
            \STATE ${X}'\leftarrow {X}'\cup\left\{{x}'\right\}$
		\ENDFOR
		\STATE  delete the $\lambda$ individuals with the lowest fitness value of ${X}'$ uniformly at random
            \STATE $X_{t+1} \leftarrow$ ${X}'$
		\STATE $t \leftarrow t+1$
		\ENDWHILE
            \STATE $x_{opt} \leftarrow$ choose a $x$ from $X_{t}$ with the maximum fitness value 
		\RETURN $x_{opt}$
	\end{algorithmic}
\end{algorithm}

In the ($\mu+\lambda$) EA, the mutation probability is conventionally set to $\frac{1}{n}$, as described in Algorithm 1. The reason for setting the mutation probability to $\frac{1}{2}$ is to streamline the analysis of expected multiple-gain. With a mutation probability of $\frac{1}{2}$, the probability of any non-best individual mutating into the global optimum is $\frac{N_{opt}}{2^{n}}$. Given that each non-best individual has an equal probability of mutating to the global optimum, Algorithm 2 accepts any offspring individuals ${x}'$ generated by the non-best individuals if $f({x}')<f(x_{t})$.

\textbf{\emph{Theorem 5: }}{Let $Y_{t}=f(x_{opt})-f(x_{t})$ and $T_{0}=min\left\{t\ge 0:Y_{t}=0\right\}$, the average-case EFHT of Algorithm 2 with mutation probability $\frac{1}{2}$  for $k$-MAX-SAT problem satisfies}
\begin{align}
	E(T_{0}|Y_{0})\le \frac{\sum_{x=1}^{s}\frac{1}{x}}{(1-e^{\frac{-\lambda N_{opt}}{2^{n}}})}. 	
\end{align}
where $N_{opt}$ denote the number of global optimums.

\emph{Proof:}\quad Assume that $Y_{t}=r$ where $t<T_{0}$. The probability $p$ that there is at least one individual among the $\lambda$ offspring individuals is the global optimum $x_{opt}$ satisfies
\begin{align*}
    p&=1-(1-\frac{N_{opt}}{2^{n}})^{\lambda}\ge 1-e^{\frac{-\lambda N_{opt}}{2^{n}}}
\end{align*}



Therefore,
\begin{align*}
	G(t,1)&=E(Y_{t}-Y_{t+1}|Y_{t}=r)\\
	&\ge p\cdot r\\
	&\ge (1-e^{\frac{-\lambda N_{opt}}{2^{n}}})\cdot r
\end{align*}

Let $h_{2}(r)=(1-e^{\frac{-\lambda N_{opt}}{2^{n}}})\cdot r$. Assume that $Y_{0}=r_{p}$, $f(x_{opt})=r_{q}\le s$, $r_{0}\ge 0$. As $h_{2}(r)$ is a monotonically non-decreasing function, by Theorem 2, we have
\begin{align*}
	E(T_{0}|Y_{0}=r_{p})&\le \frac{1}{(1-e^{\frac{-\lambda N_{opt}}{2^{n}}})}\cdot\sum_{i=1}^{p}\frac{r_{i}-r_{i-1}}{r_{i}}\\
	&\le \frac{1}{(1-e^{\frac{-\lambda N_{opt}}{2^{n}}})}\cdot\sum_{i=1}^{q}\frac{r_{i}-r_{i-1}}{r_{i}}
\end{align*}

By the inequality
\begin{align*}
	\frac{r_{i}-r_{i-1}}{r_{i}}\le \frac{1}{r_{i-1}+1}+\frac{1}{r_{i-1}+2}+\dots+\frac{1}{r_{i}}
\end{align*}

we have
\begin{align*}
	\sum_{i=1}^{q}\frac{r_{i}-r_{i-1}}{r_{i}}\le \sum_{x=r_{0}+1}^{r_{q}}\frac{1}{x} \le \sum_{x=1}^{s}\frac{1}{x}
\end{align*}

This means $E(T_{0}|Y_{0})\le \frac{1}{(1-e^{\frac{-\lambda N_{opt}}{2^{n}}})}\sum_{x=1}^{s}\frac{1}{x}$. $\hfill{\blacksquare}$

Theorem 5 provides an average-case upper bound of EFHT in closed-form expression of ($\mu+\lambda$) EA with mutation probability $\frac{1}{2}$ for any $k$-MAX-SAT instances. A thorough study of this specific class of ($\mu+\lambda$) EA with a special mutation operator will help reveal the dynamic behavior of general ($\mu+\lambda$) EA. Moreover, the conclusion of Theorem 5 can lead to the following Corollary 3.

\textbf{\emph{Corollary 3: }}{Let $Y_{t}=f(x_{opt})-f(x_{t})$ and $T_{0}=min\left\{t\ge 0:Y_{t}=0\right\}$, the EFHT of Algorithm 2 with mutation probability $\frac{1}{2}$ for $k$-MAX-SAT problem satisfies}
\begin{align}
	E(T_{0}|Y_{0}=r_{L})\in O(\frac{2^{n}(lns+1)}{\lambda N_{opt}}). 	
\end{align}
where $N_{opt}$ denote the number of global optimums.

\emph{Proof:}
According to Theorem 5, we have
\begin{align*}
	E(T_{0}|Y_{0})&\le \frac{\sum_{x=1}^{s}\frac{1}{x}}{1-e^{\frac{-\lambda N_{opt}}{2^{n}}}}
\end{align*}

According to $\sum_{x=1}^{n}\frac{1}{x}\le \ln n+1$, we have
\begin{align*}
    E(T_{0}|Y_{0})&\le \frac{lns+1}{1-e^{\frac{-\lambda N_{opt}}{2^{n}}}}
\end{align*}
When $n\rightarrow \infty$,
\begin{align*}
    \lim_{n \to \infty}\frac{lns+1}{1-e^{\frac{-\lambda N_{opt}}{2^{n}}}} &= \lim_{n \to \infty}\frac{lns+1}{1-(1-\frac{\lambda N_{opt}}{2^{n}})}\\
    &=\frac{2^{n}(lns+1)}{\lambda N_{opt}}
\end{align*}
\begin{spacing}{1}
	It means that $E(T_{0}|Y_{0}=r_{L})\in O(\frac{2^{n}(lns+1)}{\lambda N_{opt}})$. $\hfill{\blacksquare}$
\end{spacing}

By Corollary 1 and Theorem 3, we can obtain the worst-case EFHT of $(\mu+\lambda)$ EA for $k$-MAX-SAT problem.

\textbf{\emph{Theorem 6: }}{Let $Y_{t}=f(x_{opt})-f(x_{t})$ and $T_{0}=min\left\{t\ge 0:Y_{t}=0\right\}$, the worst-case EFHT of Algorithm 3 with mutation probability $\frac{1}{2}$  for $k$-MAX-SAT problem satisfies}
\begin{align}
        E(T_{0}|Y_{0})&\le\frac{\beta(Y_{0}-r_{0})}{\alpha(1-e^{\frac{-\lambda N_{opt}}{2^{n}}})}
\end{align}
where $N_{opt}$ denote the number of global optimums.

\emph{Proof: } By the proof of Theorem 5, we have $G(t, 1)\ge h_{2}(r_{p})$. Therefore, to achieve an expected multiple-gain of at least $r_{p}-r_{p-1}$ after $k$ iterations, $k$ must satisfy the condition $k\cdot h_{2}(r_{p})\ge r_{p}-r_{p-1}$. In addition, $\forall r_{i}\in S$, $0<\alpha\le r_{i}-r_{i-1}\le \beta$. By the inequality $k\cdot h_{2}(r_{p})\ge r_{p}-r_{p-1}$, we have 
\begin{align*}
	k\ge \frac{\beta}{h_{2}(r_{p})}
\end{align*}

Therefore,
\begin{align}
    k_{low}=\beta/{(1-e^{\frac{-\lambda N_{opt}}{2^{n}}})}
\end{align}

By Corollary 1, we can obtain the worst-case EFHT of Algorithm 1 for the $k$-MAX-SAT problem satisfies
\begin{align}
    E(T_{0}|Y_{0}=r_{L})&\le \frac{k_{low}(r_{L}-r_{0})}{\alpha}\\
    &=\frac{\beta(Y_{0}-r_{0})}{\alpha(1-e^{\frac{-\lambda N_{opt}}{2^{n}}})}
\end{align}
$\hfill{\blacksquare}$




 \subsection{Case Study III: Expected First Hitting Time Analysis of ($\mu+\lambda$) EA for the TSP problem where the finite point set of cities is in convex position}

The traveling salesman problem (TSP) is also a well-known combinatorial optimization problem. Nevertheless, the theoretical studies of EAs for the TSP problem remain relatively limited. Zhou \cite{zhou2009tsp} introduced a
first running time analysis of a basic ant colony optimization (ACO) algorithm on
two specific TSP problem instances. Kötzing et al. \cite{kotzing2012tsp} further proved that the ACO algorithm can obtain an approximation ratio in expected polynomial time for random TSP instances. Sutton et al. \cite{sutton2014tsp}
introduced a parameterized running-time analysis of the EAs for the Euclidean TSP. Peng et al. \cite{peng2016tsp}
showed that two simple ACO algorithms can obtain an approximation ratio in expected polynomial time for the TSP(1,2) problem. Xia et al. \cite{2019tsp} also presented an approximation running-time analysis for the TSP(1,2) problem.

Let $V=\left\{1,2,\dots,n\right\}$ be a set 
of $n$ points that denotes the cities. Let $E$ be the set of edges where $E\subseteq V\times V$. Let $G=(V,E)$ be an undirected weighted graph. Let $\left\{i,j\right\}$ denote the edge connecting city $i$ and city $j$. A candidate solution to the TSP problem is a permutation $x=\left\{x_{1},x_{2},\dots,x_{n}\right\}$ where $x_{i}=j$ denotes the city $j$ on the $i$-th position of the tour. The
Hamilton cycle in graph $G$ is the set of $n$ edges $\left\{\left\{x_{1},x_{2}\right\},\left\{x_{2},x_{3}\right\},\dots,\left\{x_{n-1},x_{n}\right\},\left\{x_{n},x_{1}\right\}\right\}$. Consequently, the goal of TSP problem is to find a permutation $x$ that forms a Hamiltonian cycle with the minimum weight.

This subsection analyzes the TSP problem where $V$ is finite a point set in convex position. A finite point set $V$ is in convex position when every point in $V$ is a vertex on its convex
hull. Deǐneko et al. \cite{dei2006tsp} noted that the TSP problem is readily solvable when $V$ is
in convex position. The optimal solution to such TSP problems is any tour that follows the boundary of the convex hull.

Let $x_{n+1}=x_{1}$. Define the function
\begin{align*}
        s(x_{i})=\begin{cases}
0, if\quad 1\le x_{i}\le n-1, |x_{i}-x_{i+1}|=1
 \\
0, if \quad  x_{i}=n,|x_{i}-x_{i+1}|\in \left\{1,n-1\right\}
\\
1, else
\end{cases}
\end{align*}
We say that the point $x_{i}$ is in the correct order if $s(x_{i})=0$. Additionally, $s(x_{i})=0$ implies that $x_{i}$ and $x_{i+1}$ are neighbors. Therefore, the objective of the TSP problem can be transformed into minimizing the number of points that are not in the correct order in the permutation $x$
\begin{align*}
    min\quad f(x)=\sum_{i=1}^{n}s(x_{i})
\end{align*}
where $f(x)\in\left\{0,2,3,\dots,n\right\}$ when $n>5$. Therefore, for $\forall r_{p}\in S$, $1=\alpha\le r_{p}-r_{p-1}\le 2=\beta$.

Since the candidate solutions of the TSP problem are no longer binary strings, mutations can no longer be performed through bit flipping. In this subsection, we consider the 2-opt inversion operation to induce mutations in the candidate solutions.

\textbf{\emph{Definition 7: }}{The 2-opt inversion operation $I_{i,j}$ transforms permutation $x$
into one another $I_{i,j}(x)$ by segment reversal \cite{sutton2014tsp}. For two positions $i$ and $j$ where $1\le i<j\le n$, the new permutation $I_{i,j}(x)$ is generated by inverting the subsequence in $x$ from $x_{i}$ to $x_{j}$.}
\begin{align*}
    x=(x_{1},\dots,x_{i-1},x_{i},x_{i+1},\dots,x_{j-1},x_{j},x_{j+1},\dots,x_{n})\\
    I_{i,j}(x)=(x_{1},\dots,x_{i-1},x_{j},x_{j-1},\dots,x_{i+1},x_{i},x_{j+1},\dots,x_{n})
\end{align*}
$\hfill{\blacksquare}$

Detailed operations of ($\mu+\lambda$) EA for the TSP problem are described as Algorithm 3 \cite{sutton2014tsp}. Algorithm 3 rejects any offspring individuals ${x}'$ generated by the non-best individuals if $f({x}')<f(x_{t})$ (lines 9-11).

\begin{algorithm}
	\caption{($\mu+\lambda$) EA for the TSP problem}
	\begin{algorithmic}[1]
		\REQUIRE {Number of cities $n$.}
		\ENSURE {The optimal permutation $x_{opt}$.}
		\STATE {Initialization: set generation $t$=0, generate a population 
			$X_{0}= \left\{x_{1}, \dots,  x_{\mu}\right\}$ with $\mu$ permutations $x_{i}=\left\{x_{i,1}, \dots,  x_{i,n}\right\}$, $1\le i\le \mu$, uniformly at random.}
		\WHILE{an optimal permutation has not been attended}
            \STATE ${X}'=X_{t}$
		\FOR{$i=1$ to $\lambda$}
            \STATE choose a $x$ from $X_{t}$ uniformly at random
            \STATE draw $s$ from Poisson distribution with parameter $\lambda_{p}$
            \STATE $x_{temp}=x$
		\STATE ${x}' \leftarrow$ perform $s+1$ random 2-opt inversion operations on $x$
            \IF{$f({x}')<f(x_{t})$ and $x_{temp}\ne x_{t}$} 
            \STATE ${x}' \leftarrow x_{temp}$
            \ENDIF
            \STATE ${X}'\leftarrow {X}'\cup\left\{{x}'\right\}$
		\ENDFOR
		\STATE delete the $\lambda$ individuals with the highest fitness value of ${X}'$ uniformly at random
            \STATE $X_{t+1} \leftarrow {X}'$
		\STATE $t \leftarrow t+1$
		\ENDWHILE
            \STATE $x_{opt} \leftarrow$ choose a $x$ from $X_{t}$ with the minimum fitness value 
		\RETURN $x_{opt}$
	\end{algorithmic}
\end{algorithm}

Theorem 7 gives the upper bound of EFHT of Algorithm 3 for the TSP problem where $V$ is a finite point set in convex position.

\textbf{\emph{Theorem 7: }}{Let $V$ be a finite set of cities in convex position where $|V|>5$. The average-case EFHT of Algorithm 3 for the TSP problem satisfies}
\begin{align}
    E(T_{0}|f(x_{0})=L)\le \frac{2}{1+g}(L+\frac{\mu e^{\lambda_{p}} C_{n}^{2}}{\lambda\lambda_{p}}\sum_{x=1}^{L}\frac{1}{x})
\end{align}
where $g=\frac{-(n-2)(n-5)+2(n-3)(n-4)}{(n-2)^{2}(n-3)}$ and $\lambda_{p}$ is the parameter of Poisson distribution in Algorithm 3.

\emph{Proof:} There are $r$ points that are not in the correct order when $f(x_{t})=r$, $t<T_{0}$. Assume that $x_{t, i}$ is one of the points not in the correct order, and $x_{t, i+1}$ is not the neighbor of $x_{t, i}$. Assume that $x_{t, j}$ is the neighbor of $x_{t, i}$. Let $E_{1}$ denote the event that the mutation operation performs only one random 2-opt inversion operation on $x_{t}$. The probability of $E_{1}$ is $\frac{\lambda_{p}}{e^{\lambda_{p}}}$. If $j>i+1>i$ or $i+1=1<j<i=n$, we say that $x_{t}$ can be improved by a single 2-opt inversion operation $I_{i+1,j}$. 
\begin{align*}
    x_{t}=(\dots,x_{i},x_{i+1},\dots,x_{j},x_{j+1},\dots)\\
    I_{i+1,j}(x_{t})=(\dots,x_{i},x_{j},\dots,x_{i+1},x_{j+1},\dots)
\end{align*}

Note that the edges $\left\{x_{i},x_{i+1}\right\}$ and $\left\{x_{j},x_{j+1}\right\}$ are transformed into $\left\{x_{i},x_{j}\right\}$ and $\left\{x_{i+1},x_{j+1}\right\}$. Therefore, $f(x_{t})$ will decrease 1 because $s(x_{i})$ will decrease from 1 to 0. Let $E_{2}$ denote the event that a random 2-opt inversion operation, which can improve $x_{t}$ is performed on $x_{t}$. The following proof will be discussed in two cases regarding $x_{j+1}$.

\textbf{Case 1}: $x_{j+1}$ is the neighbor of $x_{j}$ with the probability $\frac{1}{n-2}$. In this case, $f(x_{t})$ will increase 1 because the edge $\left\{x_{j},x_{j+1}\right\}$ is transformed into $\left\{x_{i+1},x_{j+1}\right\}$. Additionally, if $x_{i+1}$ is the neighbor of $x_{j+1}$ with the probability $\frac{1}{n-3}$, $f(x_{t})$ will decrease 1. Therefore, the variation of $f(x_{t})$ is
\begin{align*}
    \Delta(f(x_{t}))=-1+\frac{1-\frac{1}{n-3}}{(n-2)}
    &=-1+\frac{(n-4)}{(n-2)(n-3)}
\end{align*}

\textbf{Case 2}: $x_{j+1}$ is not the neighbor of $x_{j}$ with the probability $1-\frac{1}{n-2}$. If $x_{j+1}$ is the neighbor of $x_{i}$ and $x_{i+1}$ is the neighbor of $x_{j+1}$ with the probability $\frac{1}{(n-3)^{2}}$, $f(x_{t})$ will decrease 1. Additionally, if $x_{j+1}$ is not the neighbor of $x_{i}$ and $x_{i+1}$ is the neighbor of $x_{j+1}$ with the probability $(1-\frac{1}{n-3})\cdot\frac{2}{(n-2)}$, $f(x_{t})$ will decrease 1. Therefore, the variation of $f(x_{t})$ is
\begin{align*}
    \Delta(f(x_{t}))&=-1+\frac{\frac{-1}{(n-3)^{2}}-\frac{2(n-4)}{(n-2)(n-3)}}{\frac{n-2}{n-3}}\\
    &=-1-\frac{(n-2)+2(n-3)(n-4)}{(n-2)^{2}(n-3)}
\end{align*}

Let $g=-\frac{(n-4)}{(n-2)(n-3)}+\frac{(n-2)+2(n-3)(n-4)}{(n-2)^{2}(n-3)}$. Therefore,
\begin{align*}
    E(f(x_{t})-f(I(x_{t}))|E_{1}\cap E_{2})&=1+g
\end{align*}

Since the number of all possible 2-opt inversion operation is $C_{n}^{2}$, we have the probability of $E_{2}$ is at least $\frac{r}{C_{n}^{2}}$. The probability of creating an offspring permutation $x$ by $x_{t}$ where $f(x)<f(x_{t})$ is at least $\frac{b_{t}}{\mu}\frac{\lambda_{p}}{e^{\lambda_{p}}}\frac{r}{C_{n}^{2}}$. The probability that this event occurs at least once during $\lambda$ generations is 
\begin{align*}
    p&=1-(1-\frac{b_{t}}{\mu}\frac{\lambda_{p}}{e^{\lambda_{p}}}\frac{r}{C_{n}^{2}})^{\lambda}\\
    &\ge1-\frac{1}{1+\lambda\frac{b_{t}}{\mu}\frac{\lambda_{p}}{e^{\lambda_{p}}}\frac{r}{C_{n}^{2}}}\\
    &\ge\frac{1}{1+\frac{\mu e^{\lambda_{p}} C_{n}^{2}}{\lambda\lambda_{p} r}}
\end{align*}

Therefore,
\begin{align*}
    E(f(x_{t})-f(x_{t+1})|f(x_{t})=r)&\ge p(1+g)\\
    &\ge \frac{1+g}{1+\frac{\mu e^{\lambda_{p}} C_{n}^{2}}{\lambda\lambda_{p} r}}
\end{align*}

Let $h_{3}(r)=(1+g)/(1+\frac{\mu e^{\lambda_{p}} C_{n}^{2}}{\lambda\lambda_{p} r})$. Note that the range of fitness value $S=\left\{0,1,\dots,n\right\}$. As $h_{3}(r)$ is a monotonically non-decreasing function, by Theorem 2, we have
\begin{align*}
    E(T_{0}|f(x_{0})=L)&\le \frac{1}{1+g}\sum_{i=1}^{L}(r_{i}-r_{i-1})(1+\frac{\mu e^{\lambda_{p}} C_{n}^{2}}{\lambda\lambda_{p} r_{i}})\\
    &\le\frac{2}{1+g}\sum_{i=1}^{L}(1+\frac{\mu e^{\lambda_{p}} C_{n}^{2}}{\lambda\lambda_{p} r_{i}})\\
    &=\frac{2}{1+g}(L+\frac{\mu e^{\lambda_{p}} C_{n}^{2}}{\lambda\lambda_{p}}\sum_{i=1}^{L}\frac{1}{r_{i}})\\
    &=\frac{2}{1+g}(L+\frac{\mu e^{\lambda_{p}} C_{n}^{2}}{\lambda\lambda_{p}}\sum_{x=1}^{L}\frac{1}{x})
\end{align*}
where $g=\frac{-(n-2)(n-5)+2(n-3)(n-4)}{(n-2)^{2}(n-3)}$.
$\hfill{\blacksquare}$

Moreover, the conclusion of Theorem 7 can lead to the following Corollary 4.

\textbf{\emph{Corollary 4: }}{Let $V$ be a finite set of cities in convex position where $|V|>5$. The EFHT of Algorithm 3 for the TSP problem satisfies}
\begin{align}
    E(T_{0}|f(x_{0})=L)\in O(\frac{\mu}{\lambda}n^{2}\ln n+n)
\end{align}
\emph{Proof:}According to Theorem 7, we have
\begin{align*}
    E(T_{0}|f(x_{0})=L)&\le \frac{2}{1+g}(L+\frac{\mu e^{\lambda_{p}} C_{n}^{2}}{\lambda\lambda_{p}}\sum_{x=1}^{L}\frac{1}{x})\\
    &\le \frac{2}{1+g}(n+\frac{\mu e^{\lambda_{p}} C_{n}^{2}}{\lambda\lambda_{p}}\sum_{x=1}^{n}\frac{1}{x})\\
    &\le \frac{2}{1+g}(n+\frac{\mu e^{\lambda_{p}} C_{n}^{2}(\ln n+1)}{\lambda\lambda_{p}})
\end{align*}
When $n\rightarrow \infty$,
\begin{align*}
    \lim_{n \to \infty}1+g&=1+\lim_{n \to \infty}\frac{-(n-2)(n-5)+2(n-3)(n-4)}{(n-2)^{2}(n-3)}\\
    &=1
\end{align*}

Therefore, 
\begin{align*}
    \lim_{n \to \infty}E(T_{0}|f(x_{0})=L)&\le \lim_{n \to \infty}\frac{2(n+\frac{\mu e^{\lambda_{p}} C_{n}^{2}(\ln n+1)}{\lambda\lambda_{p}})}{1+g}\\
    &=2n+\frac{\mu e^{\lambda_{p}}}{\lambda\lambda_{p}}n(n-1)(\ln n+1)
\end{align*}

It means that $E(T_{0}|f(x_{0})=L)\in O(\frac{\mu}{\lambda}n^{2}\ln n+n)$. $\hfill{\blacksquare}$

Corollary 4 presents a tighter upper bound $O(\frac{\mu}{\lambda}n^{2}\ln n+n)$ than the result $O(n\cdot max\left\{(\mu/\lambda)n^{2},1\right\})$ in \cite{sutton2014tsp} by a factor of $n/\ln n$ when $(\mu/\lambda)n^{2}>1$ and $n>5$.

By Corollary 1 and Theorem 7, we can also obtain the worst-case EFHT of Algorithm 3 for TSP problem.

\textbf{\emph{Theorem 8: }}{Let $V$ be a finite set of cities in convex position where $|V|>5$. The worst-case EFHT of Algorithm 3 for the TSP problem satisfies}
\begin{align}
    E(T_{0}|f(x_{0})=L)&\le \frac{\beta(L-r_{0})(1+\frac{\mu e^{\lambda_{p}} C_{n}^{2}}{\lambda\lambda_{p}})}{\alpha(1+g)}
\end{align}
where $g=\frac{-(n-2)(n-5)+2(n-3)(n-4)}{(n-2)^{2}(n-3)}$ and $\lambda_{p}$ is the parameter of Poisson distribution in Algorithm 3.

\emph{Proof: } By the proof of Theorem 7, we have $G(t, 1)\ge h_{3}(r_{p})$. Therefore, to achieve an expected multiple-gain of at least $r_{p}-r_{p-1}$ after $k$ iterations, $k$ must satisfy the condition $k\cdot h_{3}(r_{p})\ge r_{p}-r_{p-1}$. In addition, $\forall r_{i}\in S$, $0<\alpha\le r_{i}-r_{i-1}\le \beta$. By the inequality $k\cdot h_{3}(r_{p})\ge r_{p}-r_{p-1}$, we have 
\begin{align*}
	k&\ge \frac{\beta}{h_{3}(r_{p})}
    =\frac{\beta(1+\frac{\mu e^{\lambda_{p}} C_{n}^{2}}{\lambda\lambda_{p} r})}{1+g} 
\end{align*}

Therefore,
\begin{align}
    k_{low}=\beta(1+\frac{\mu e^{\lambda_{p}} C_{n}^{2}}{\lambda\lambda_{p}})/(1+g)
\end{align}

By Corollary 1, we can obtain the worst-case EFHT of Algorithm 3 for the TSP problem satisfies
\begin{align}
    E(T_{0}|f(x_{0})=L)&\le \frac{k_{low}(L-r_{0})}{\alpha}\\
    &=\frac{\beta(L-r_{0})(1+\frac{\mu e^{\lambda_{p}} C_{n}^{2}}{\lambda\lambda_{p}})}{\alpha(1+g)}
\end{align}
$\hfill{\blacksquare}$







\section{Experiments}
We conduct the following three experiments to verify whether the experimental results align with the theorems in Section IV. The experiments include ($\mu+\lambda$) EA for the knapsack problem with favorably correlated weights, ($\mu+\lambda$) EA for the $k$-MAX-SAT problem, and ($\mu+\lambda$) EA for the TSP problem where the finite point set of cities is in convex position.

\subsection{Experimental Setting}
	In the following three experiments, we ran the algorithm 1000 times independently for each encoding length $n$ following references \cite{bian2022better,bian2023stochastic,bian2024archive,lu2024towards}. Let $T_{i}$ denote the FHT of the $i$-th run, and $T_{max}=max\left\{T_{i}|1\le i\le1000\right\}$ denote the largest FHT among 1000 runs. $\widehat{T_{0}}=\sum_{i=1}^{1000}T_{i}/{1000}$ is considered to be the estimation of the actual
	EFHT. Let $k_{i}$ denote the most prolonged time interval during which the gain remains zero in the $i$-th run. $\widehat{k}=\sum_{i=1}^{1000}k_{i}/{1000}$ was considered to be the estimation of $k_{low}$. Let $EFHT_{average}$ denote the theoretical average-case upper bound of EFHT. Let $EFHT_{worst}$ denote the theoretical worst-case upper bound of EFHT.

The correlation coefficient is commonly used to describe the relationship between variables. Therefore, we utilize the correlation coefficient to assess whether the experimental results align with the theorems in Section III. Formula (25) \cite{1997Statistics} is to calculate the correlation coefficient r(x,y) as follows.
\begin{align}
	r(x,y)=\frac{n\sum xy-\sum x\sum y}{\sqrt{n\sum x^2-\left(\sum x\right)^2}\sqrt{n\sum y^2-\left(\sum y\right)^2}}
\end{align}

As $|r(x,y)|$ approaches 1, it indicates a stronger correlation between $x$ and $y$, while $|r(x,y)|$ approaching to $0$ signifies a weaker correlation. For each experiment, we will present the computed results for the three correlation coefficients: $r(EFHT_{average},T_{0})$, $r(EFHT_{worst},T_{max})$, and $r(\widehat{k},k_{low})$. The criteria for evaluating the correlation coefficient may vary among researchers from different fields. However, a correlation coefficient greater than 0.91 generally indicates a strong relationship\cite{1997Statistics}. Therefore, if $EFHT_{average}>\widehat{T_{0}}$, $EFHT_{worst}>T_{max}$, $\widehat{k}>k_{low}$, $r(EFHT_{average},T_{0})>0.91$, $r(EFHT_{worst},T_{max})>0.91$, and $r(\widehat{k},k_{low})>0.91$, the experimental results align with the theorems in Section IV.

We used self-constructed cases rather than the test dataset to simplify the calculation. Because we are using self-constructed cases, there is no need to estimate the value of $\alpha$. In addition, we implemented Algorithms 1 and 2 using Matlab. 

\subsection{Verification of Theorems 3 and 4}
The parameters of Algorithm 1 for the knapsack problem with favorably correlated weights are set as follows: the parent
population size is set to $\mu=2$, the offspring population size is set to $\lambda=10$, the capacity of knapsack is set to $K=3$, the values of the first three items are set to 3, 3, 1 with weights of 1 each, the values of all the remaining items are set to 1 with weights of 2 each, the initial population is set to $X_{0}=\left\{x_{0}, x_{0}, \dots, x_{0}\right\}$ where $x_{0}=\left\{0,0,\dots,0\right\}$, and the encoding length $n$ is chosen from the set $\left\{20,21,22,\dots,40\right\}$. 

\begin{figure*}
	\centering
	\subfloat[\label{a}]{
		\includegraphics[width=0.28\textwidth]{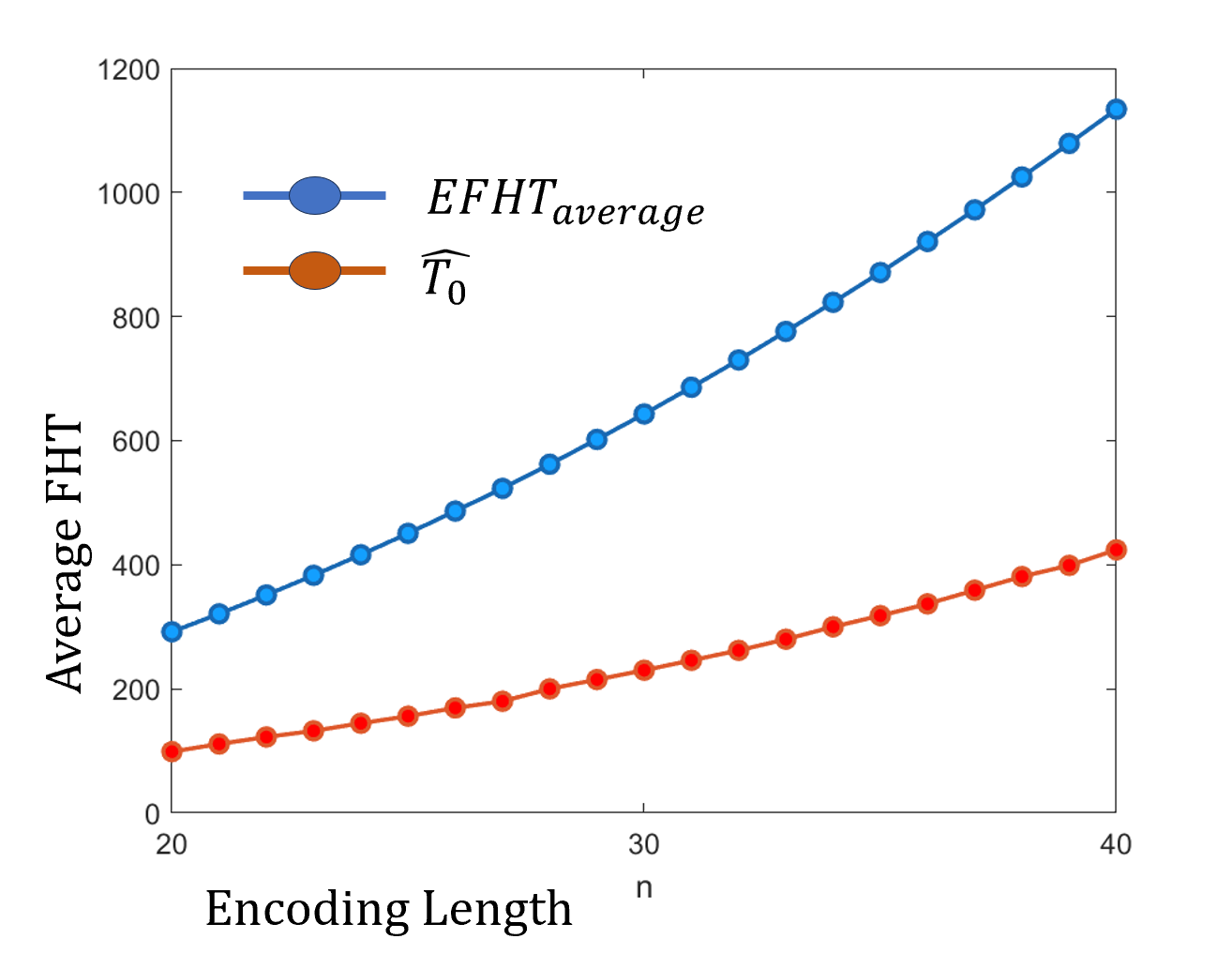}}
	\subfloat[\label{b}]{
		\includegraphics[width=0.28\textwidth]{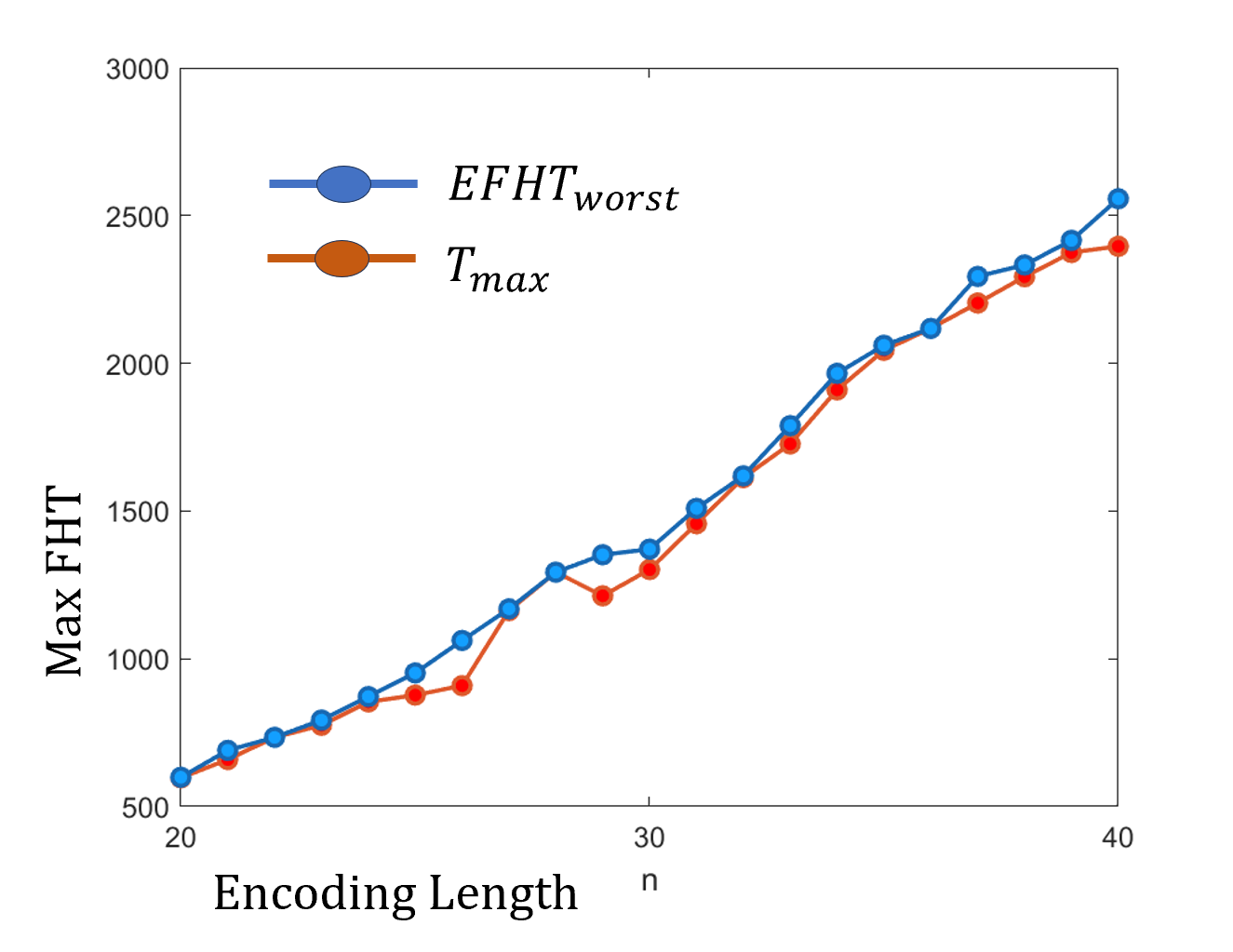}}
	\subfloat[\label{c}]{
		\includegraphics[width=0.28\textwidth]{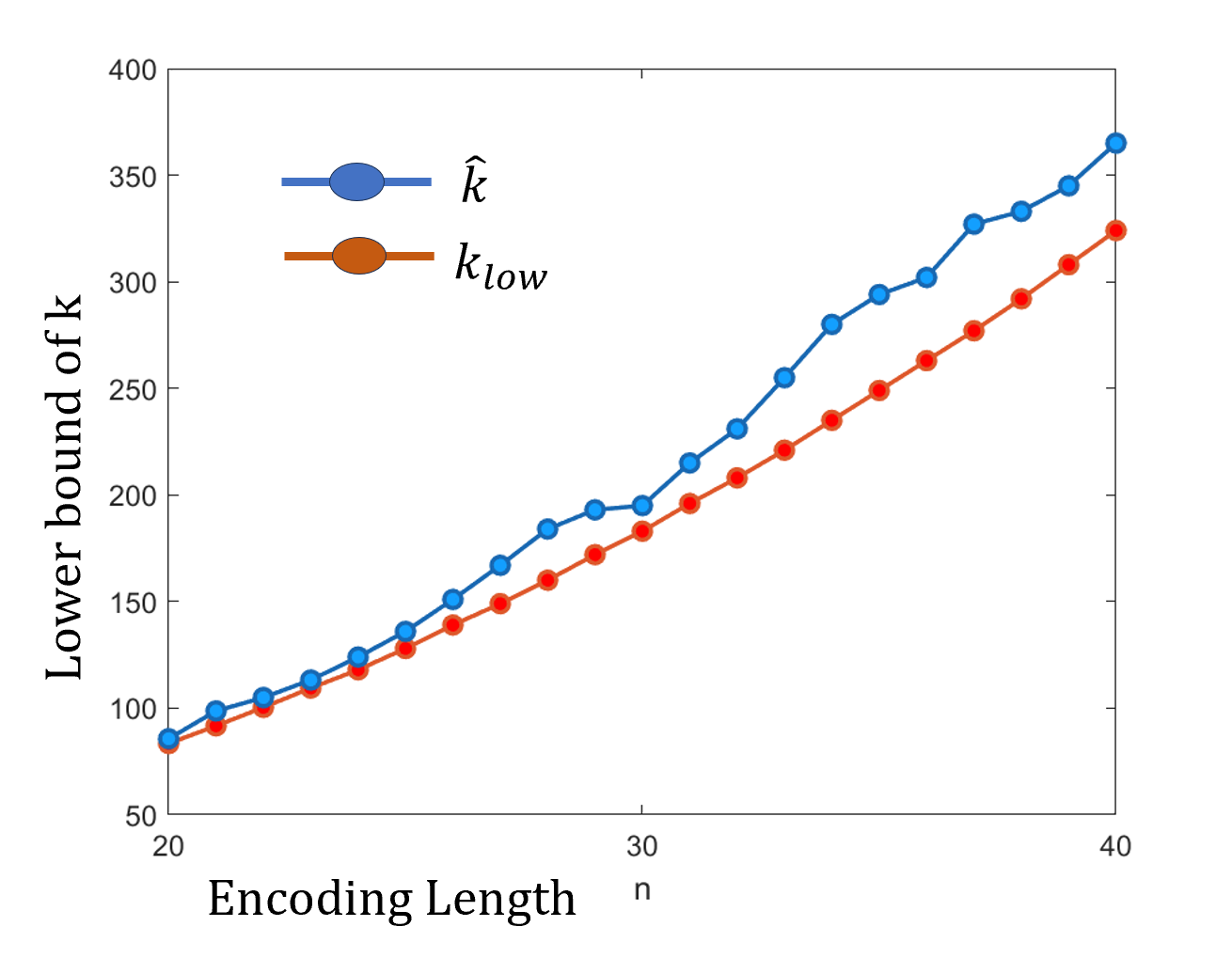}}
	\caption{Results of Algorithm 1 for the knapsack problem with favorably correlated weights. (a) Theoretical average-case upper bounds of EFHT given by Formula (7) and estimation of actual EFHT $\widehat{T_{0}}=\sum_{i=1}^{1000}T_{i}/{1000}$. (b) Theoretical worst-case upper bounds of EFHT given by Formula (11) and actual largest FHT $T_{max}=max\left\{T_{i}|1\le i\le1000\right\}$. (c) Estimation of $k_{low}$ given by $\widehat{k}=\sum_{i=1}^{1000}k_{i}/{1000}$ and theoretical value of $k_{low}$ given by Formula (10).}
\end{figure*}

\begin{figure*}
	\centering
	\subfloat[\label{fig:a}]{
		\includegraphics[width=0.28\textwidth]{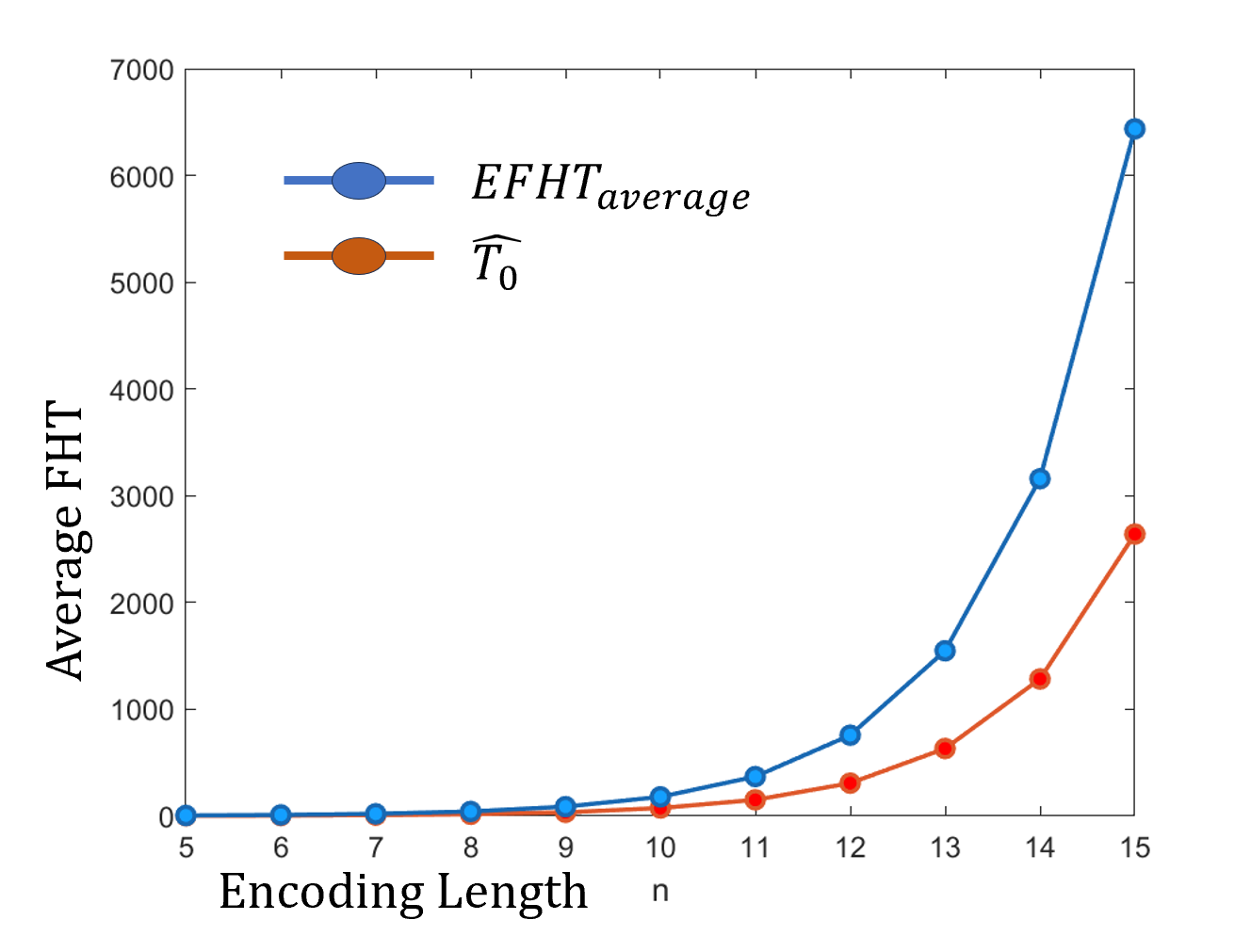}}
	\subfloat[\label{fig:b}]{
		\includegraphics[width=0.28\textwidth]{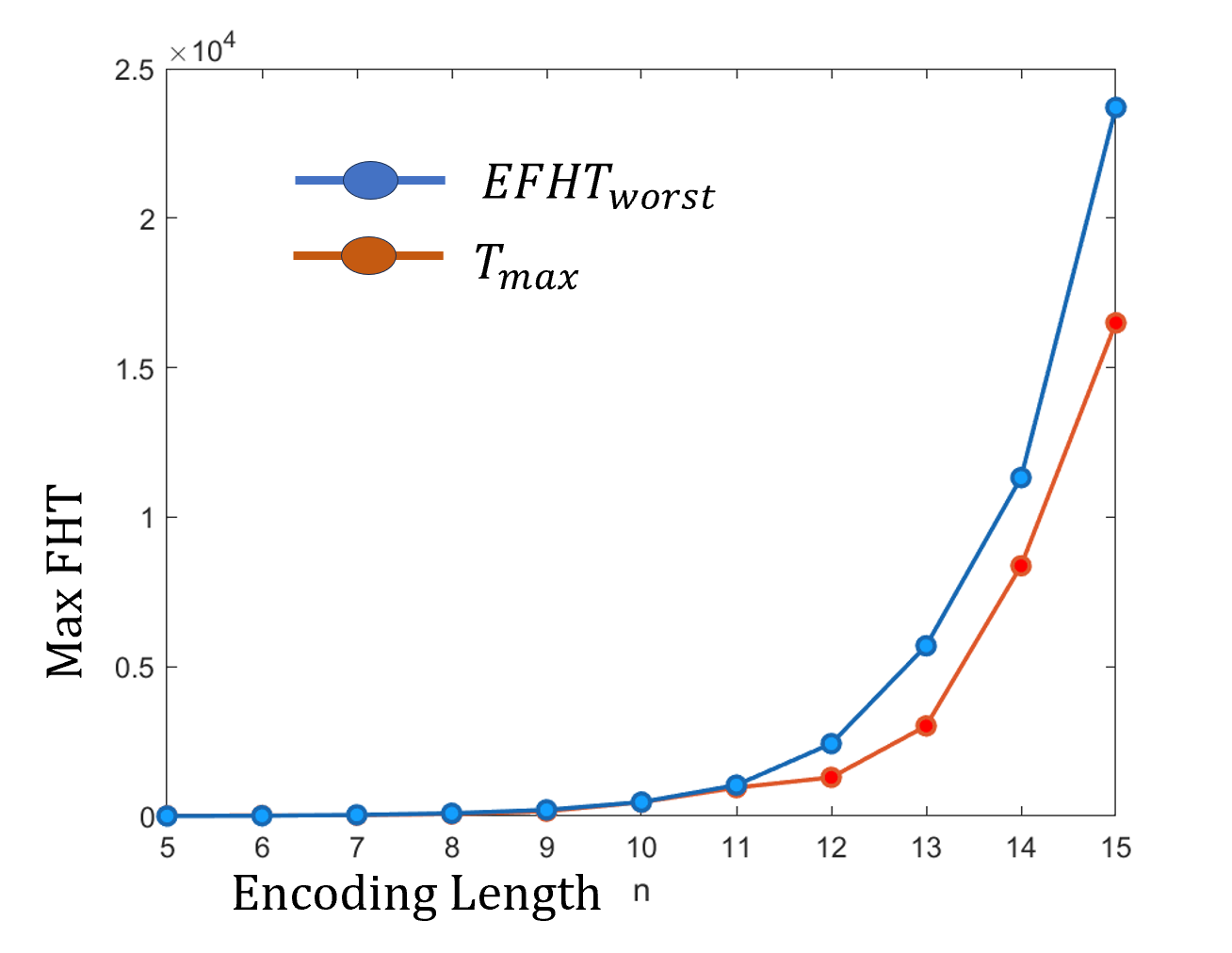}}
	\subfloat[\label{fig:c}]{
		\includegraphics[width=0.28\textwidth]{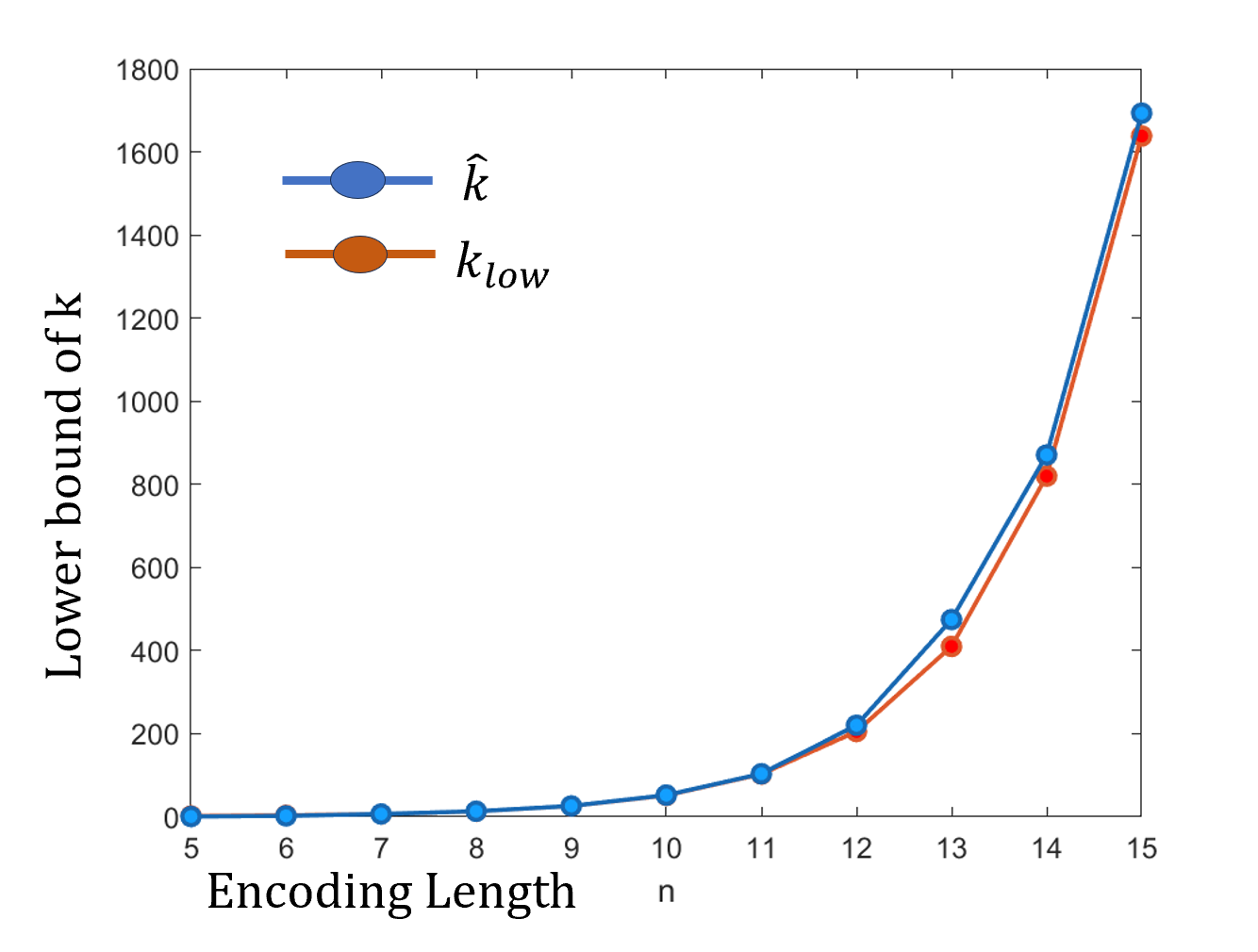}}
	\caption{Results of Algorithm 2 for the $k$-MAX-SAT problem. (a) Theoretical average-case upper bounds of EFHT given by Formula (13) and estimation of actual EFHT $\widehat{T_{0}}=\sum_{i=1}^{1000}T_{i}/{1000}$. (b) Theoretical worst-case upper bounds of EFHT given by Formula (17) and actual largest FHT $T_{max}=max\left\{T_{i}|1\le i\le1000\right\}$. (c) Estimation of $k_{low}$ given by $\widehat{k}=\sum_{i=1}^{1000}k_{i}/{1000}$ and theoretical value of $k_{low}$ given by Formula (16).}
\end{figure*}

\begin{figure*}
	\centering
	\subfloat[\label{a}]{
		\includegraphics[width=0.28\textwidth]{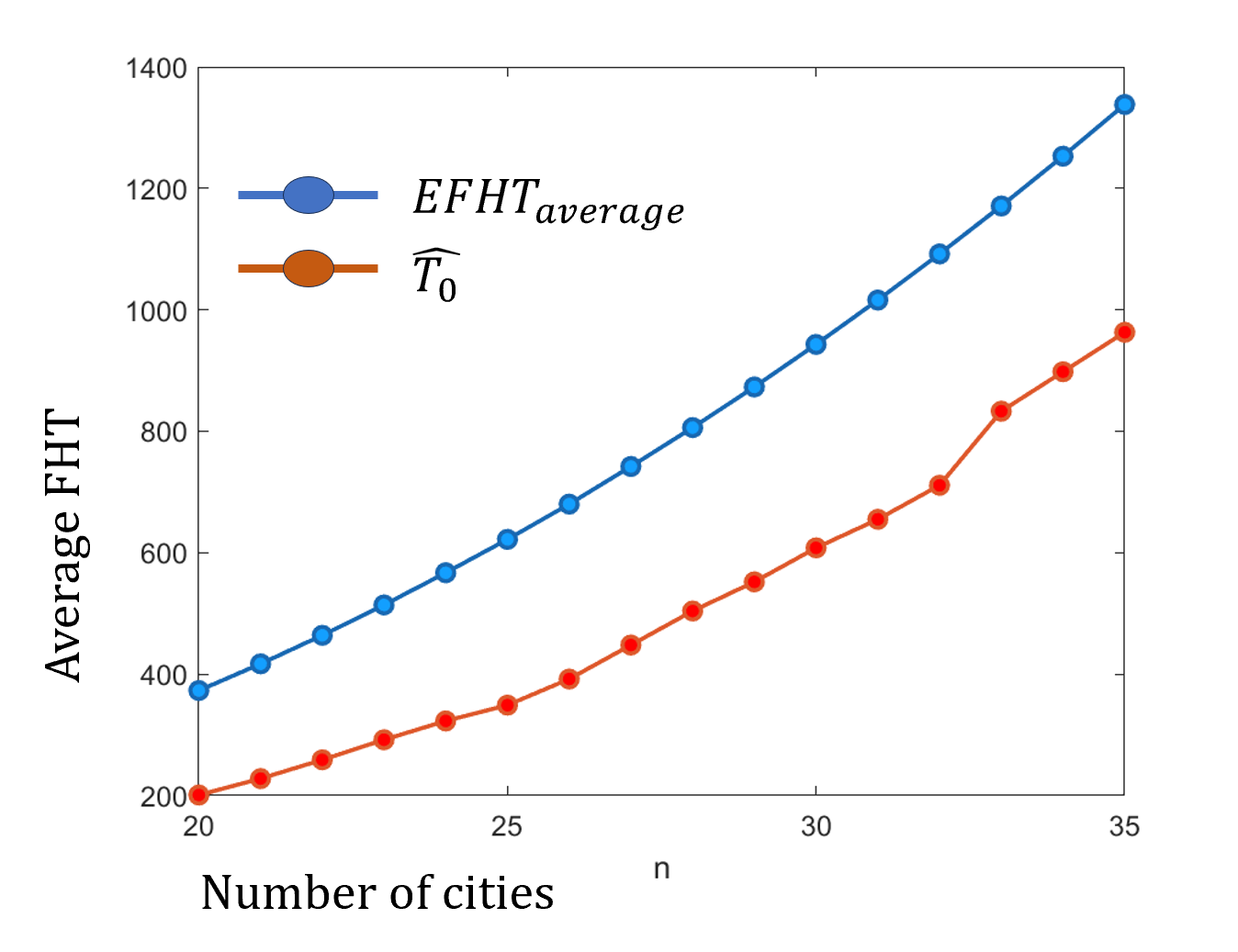}}
	\subfloat[\label{b}]{
		\includegraphics[width=0.28\textwidth]{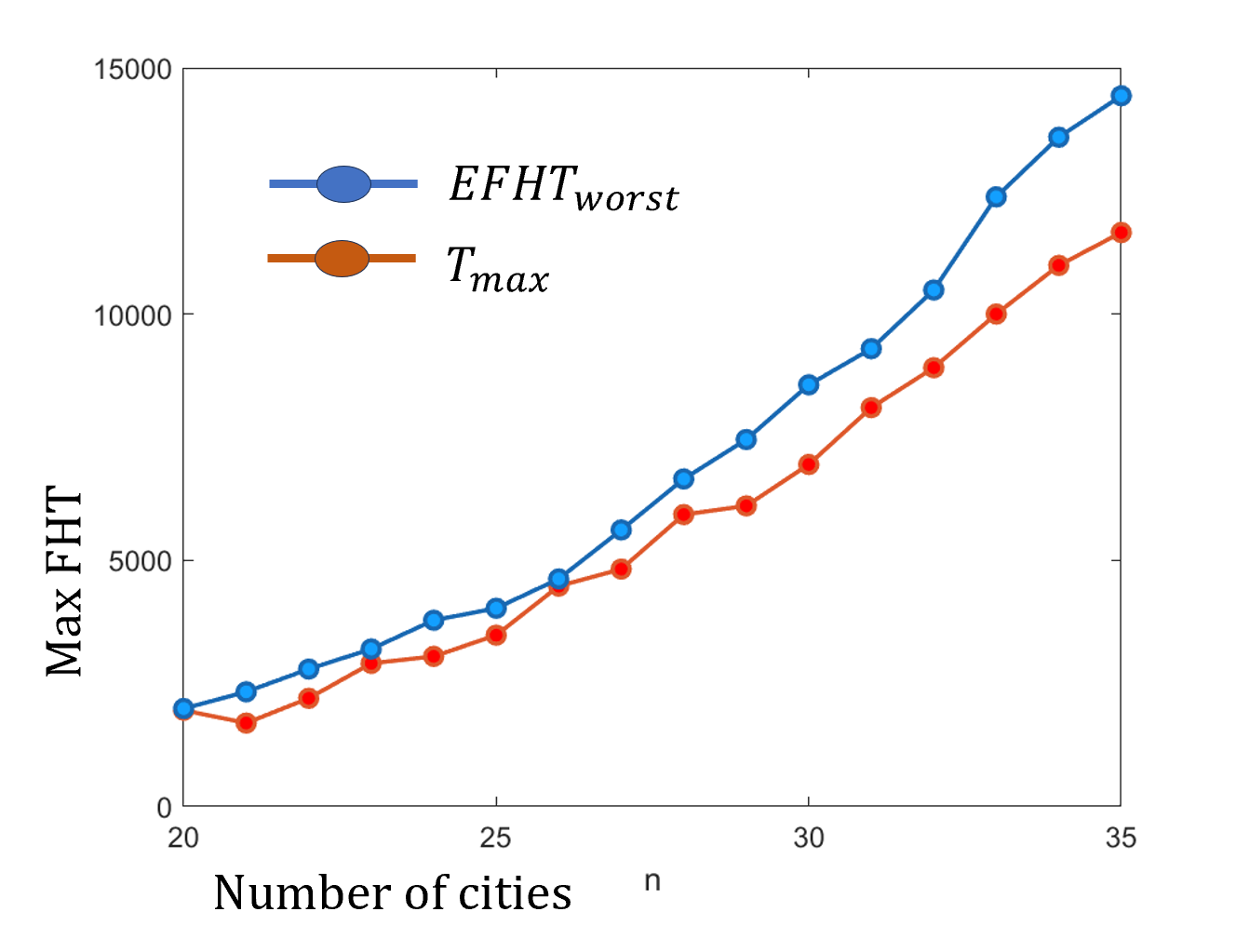}}
	\subfloat[\label{c}]{
		\includegraphics[width=0.28\textwidth]{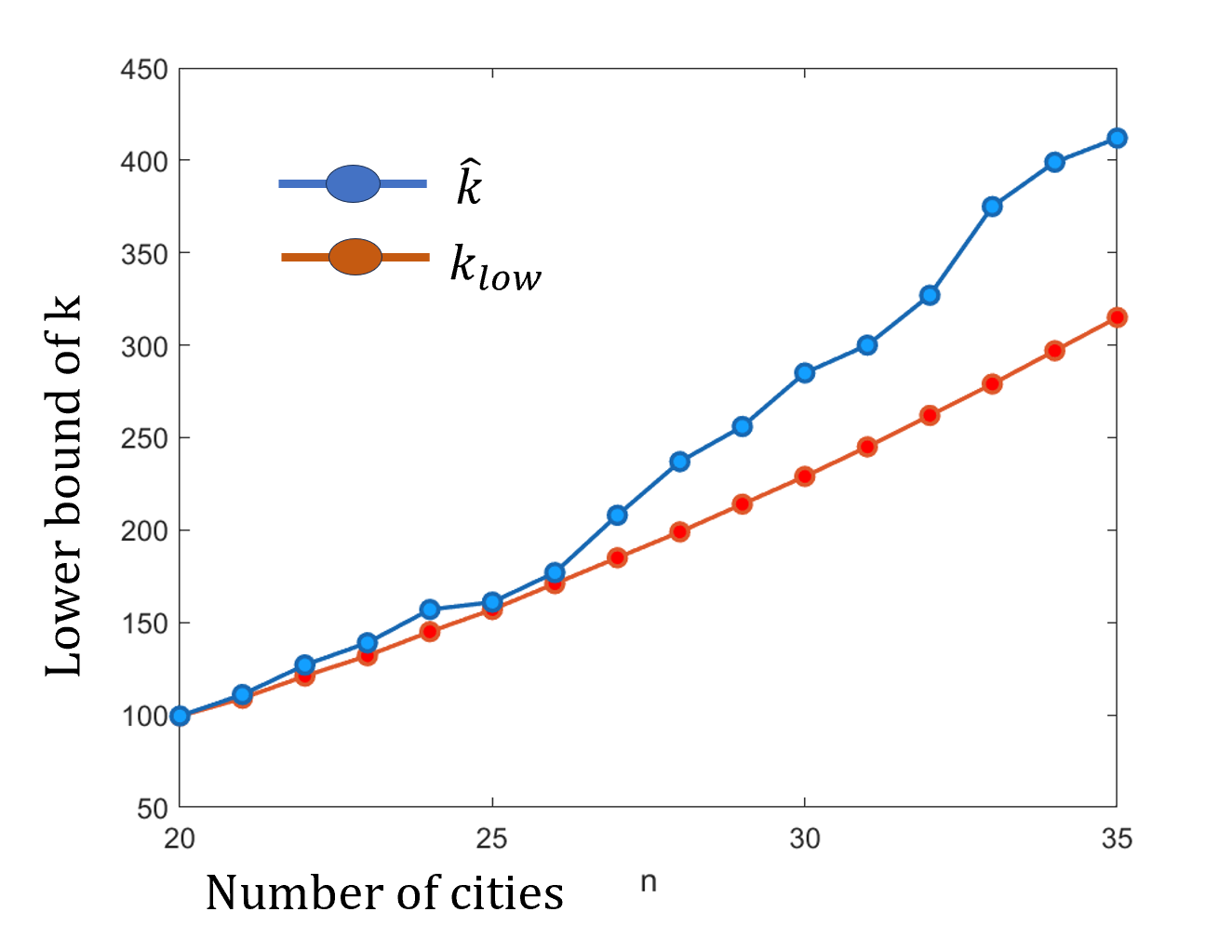}}
	\caption{Results of Algorithm 3 for the TSP problem. (a) Theoretical average-case upper bounds of EFHT given by Formula (19) and estimation of actual EFHT $\widehat{T_{0}}=\sum_{i=1}^{1000}T_{i}/{1000}$. (b) Theoretical worst-case upper bounds of EFHT given by Formula (23) and actual largest FHT $T_{max}=max\left\{T_{i}|1\le i\le1000\right\}$. (c) Estimation of $k_{low}$ given by $\widehat{k}=\sum_{i=1}^{1000}k_{i}/{1000}$ and theoretical value of $k_{low}$ given by Formula (22).}
\end{figure*}

Fig. 2(a) provides the data points of $EFHT_{average}$ given by Formula (7) and $\widehat{T_{0}}$ with respect to the encoding length.
Fig. 2(b) shows $EFHT_{worst}$ given by Formula (11) and $T_{max}$ with respect to the encoding length. Here, we replace Formula (10) with $\widehat{k}=\sum_{i=1}^{1000}k_{i}/{1000}$ when calculating Formula (11). In the subsequent two experiments, we apply the same opration when calculating $EFHT_{worst}$. 
Fig. 2(c) plots the data points of $\widehat{k}$ and $k_{low}$ given by Formula (10) with respect to the encoding length.

Table I shows the results of correlation coefficient of Algorithm 1 for the knapsack problem with favorably correlated weights. As shown in Fig. 2 and Table I, the experimental results are consistent with Theorems 3 and 4.

\begin{table}[htbp]
\centering
\small
\caption{Results of correlation coefficient of Algorithm 1 for the knapsack problem with favorably correlated weights}
\label{tab:correlation_results}
\begin{threeparttable}
\begin{tabular}{lcc}
\toprule
 & Correlation Coefficient & Consistency\tnote{1} \\
\midrule
$r(EFHT_{average},\widehat{T_{0}})$  & 0.9997 & consistent \\
$r(EFHT_{worst},T_{max})$ & 0.9969 & consistent \\
$r(\widehat{k},k_{low})$ & 0.9975 & consistent \\
\bottomrule
\end{tabular}
\begin{tablenotes}    
        \footnotesize               
        \item[1] The consistency is judged by comparing variables $x$ and $y$. If $x>y$ and $r(x,y)>0.91$, the experimental results are consistent with the theorems.
\end{tablenotes}            
\end{threeparttable}
\end{table}


\subsection{Verification of Theorems 5 and 6}
The parameters of Algorithm 2 for $k$-MAX-SAT problem are set as follows: the parent population size is set to $\mu=2$, the offspring population size is set to $\lambda=10$, the problem instance is set to $f(X)=(x_{1}\vee \neg x_{2} )\wedge\dots(x_{1}\vee \neg x_{n})\wedge(\neg x_{1}\vee x_{2} )\wedge\dots(\neg x_{1}\vee x_{n} )$, the initial population is set to $X_{0}=\left\{x_{0}, x_{0}, \dots, x_{0}\right\}$ where $x_{0}=\left\{0,1,\dots,1\right\}$, and the encoding length $n$ is chosen from the set $\left\{5,6,\dots,14,15\right\}$.

Fig. 3(a) shows the data points of $EFHT_{average}$ given by Formula (13) and $\widehat{T_{0}}$ with respect to the encoding length. 
Fig. 3(b) presents $EFHT_{worst}$ given by Formula (17) and $T_{max}$ with respect to the encoding length.
Fig. 3(c) plots the data points of $\widehat{k}$ and $k_{low}$ given by Formula (16) with respect to the encoding length.

The results of correlation coefficient of Algorithm 1 for the $k$-MAX-SAT problem are presented in Table II. According to Fig. 3 and Table II, the experimental results are consistent with Theorems 5 and 6.

\begin{table}[htbp]
\centering
\small
\caption{Results of correlation coefficient of Algorithm 1 for the $k$-MAX-SAT problem}
\label{tab:correlation_results}
\begin{threeparttable}
\begin{tabular}{lcc}
\toprule
 & Correlation Coefficient & Consistency\tnote{1} \\
\midrule
$r(EFHT_{average},\widehat{T_{0}})$  & 0.9999 & consistent \\
$r(EFHT_{worst},T_{max})$ & 0.9975 & consistent \\
$r(\widehat{k},k_{low})$ & 0.9995 & consistent \\
\bottomrule
\end{tabular}
\begin{tablenotes}    
        \footnotesize               
        \item[1] The consistency is judged by comparing variables $x$ and $y$. If $x>y$ and $r(x,y)>0.91$, the experimental results are consistent with the theorems.        
\end{tablenotes}            
\end{threeparttable}
\end{table}


\subsection{Verification of Theorems 7 and 8}
The parameters of Algorithm 3 for the TSP problem where the finite point set of cities is in convex position are set as follows: the parent population size is set to $\mu=2$, the offspring population size is set to $\lambda=10$, the parameter of Poisson distribution is set to $\lambda_{p}=1$, the number of cities $n$ is chosen from the set $\left\{20,21,,\dots,35\right\}$, and the initial population is set to $X_{0}=\left\{x_{0}, x_{0}, \dots, x_{0}\right\}$ where $x_{0}=\left\{1,3,\dots,n,2,4,\dots n-5,n-1,n-3\right\}$ when $n$ is odd ($x_{0}=\left\{1,3,\dots,n-1,2,4,\dots n-4,n,n-2\right\}$ when $n$ is even).

Fig. 4(a) shows the results of $EFHT_{average}$ given by Formula (19) and $\widehat{T_{0}}$ with respect to the number of cities. 
Fig. 4(b) provides the data points of $EFHT_{worst}$ given by Formula (23) and $T_{max}$ with respect to the number of cities. 
Fig. 4(c) shows $\widehat{k}$ and $k_{low}$ given by Formula (22) with respect to the number of cities.

Table III presents the results of correlation coefficient of Algorithm 3 for the TSP problem where the finite point set of cities is in convex position. From Fig. 4 and Table III, it can be seen that the experimental results align with Theorems 7 and 8.


\begin{table}[htbp]
\centering
\small
\caption{Results of the correlation coefficient of Algorithm 3 for the TSP problem where the finite point set of cities is in convex position}
\label{tab:correlation_results}
\begin{threeparttable}
\begin{tabular}{lcc}
\toprule
 & Correlation Coefficient & Consistency\tnote{1} \\
\midrule
$r(EFHT_{average},\widehat{T_{0}})$  & 0.9959 & consistent \\
$r(EFHT_{worst},T_{max})$ & 0.9967 & consistent \\
$r(\widehat{k},k_{low})$ & 0.9956 & consistent \\
\bottomrule
\end{tabular}
\begin{tablenotes}    
        \footnotesize               
        \item[1] The consistency is judged by comparing variables $x$ and $y$. If $x>y$ and $r(x,y)>0.91$, the experimental results are consistent with the theorems.        
\end{tablenotes}            
\end{threeparttable}
\end{table}


\section{Conclusion}
This paper proposes a multiple-gain model to estimate the EFHT of EAs. The multiple-gain model extends the concept of average gain to provide two formulas (Formula (3) and Formula (4)) to estimate the upper bounds of EFHT in both worst-case and average-case scenarios for a stochastic process. The cores of the proposed model are Theorem 1, Corollary 1 and Theorem 2. Theorem 1 presents the worst-case upper bound of EFHT for a stochastic process. Based on the Theorem 1, Corollary 1 further provides the least worst-case upper bound of EFHT. When applying Corollary 1, experimental data can replace theoretical calculations to estimate the worst-case EFHT. Theorem 2 presents the average-case upper bound of EFHT for a stochastic process.

We utilize the multiple-gain model to estimate the average-case and worst-case upper bounds of EFHT for $(\mu+\lambda)$ EA on three different combinatorial optimization problems. The estimated results are consistent
with the experimental findings, which demonstrates that the
multiple-gain model serves as an effective method for running-time analysis of $(\mu+\lambda)$ EA.

Although we analyze different instances of evolutionary combinatorial optimization in this paper, the $(\mu+\lambda)$ EA involved is still a basic population-based EAs. Future work will investigate the application of the multiple-gain model for the running-time analysis of advanced EAs in practice.

\bibliography{Multiple-gain}
\bibliographystyle{IEEEtran}

\end{document}